\documentclass[journal]{IEEEtran}
\usepackage{amsmath,amsfonts,amssymb}
\usepackage{algorithmic}
\usepackage{algorithm}
\usepackage{array}
\usepackage[caption=false,font=normalsize,labelfont=sf,textfont=sf]{subfig}
\usepackage{textcomp}
\usepackage{stfloats}
\usepackage{url}
\usepackage{verbatim}
\usepackage{graphicx}
\usepackage{cite}
\usepackage{doi}
\usepackage{booktabs}
\usepackage{hyperref}
\usepackage{color}
\usepackage{multirow}
\usepackage[table]{xcolor}
\usepackage{pdfpages}
\begin{document}

\title{Aligning Few-Step Diffusion Models with Dense Reward Difference Learning}

\author{
Ziyi Zhang,
Li Shen,
Sen Zhang,
Deheng Ye,
Yong Luo*,
Miaojing Shi,
Dongjing Shan*,
\\
Bo Du*,~\IEEEmembership{Senior Member,~IEEE,}
and Dacheng Tao,~\IEEEmembership{Fellow,~IEEE}

\thanks{Ziyi Zhang, Yong Luo, and Bo Du are with the School of Computer Science, National Engineering Research Center for Multimedia Software and Hubei Key Laboratory of Multimedia and Network Communication Engineering, Wuhan University, Wuhan 430072, China (e-mail: ziyizhang27@whu.edu.cn; yluo180@gmail.com; dubo@whu.edu.cn).}
\thanks{Li Shen is with the School of Cyber Science and Technology, Shenzhen Campus of Sun Yat-sen University, Shenzhen 518107, China (e-mail: mathshenli@gmail.com).}
\thanks{Sen Zhang is with TikTok, ByteDance, Sydney, NSW 2000, Australia (e-mail: senzhang.thu10@gmail.com).}
\thanks{Deheng Ye is with Tencent Inc., Shenzhen 518052, China (e-mail: dericye@tencent.com).}
\thanks{Miaojing Shi is with the College of Electronic and Information Engineering, Tongji University, Shanghai 201804, China (e-mail: mshi@tongji.edu.cn).}
\thanks{Dongjing Shan is with the School of Medical Information and Engineering, Southwest Medical University, Luzhou 646000, Sichuan, China (e-mail: shandongjing@swmu.edu.cn).}
\thanks{Dacheng Tao is with the College of Computing and Data Science and the Generative AI Lab at Nanyang Technological University, 50 Nanyang Avenue, Singapore 639798 (e-mail: dacheng.tao@gmail.com).}
\thanks{
\textit{*Corresponding authors: Yong Luo; Dongjing Shan; Bo Du.}
}
}

\markboth{IEEE Transactions on Pattern Analysis and Machine Intelligence,~2026}%
{Zhang \MakeLowercase{\textit{et al.}}: Aligning Few-Step Diffusion Models with Dense Reward Difference Learning}

\IEEEpubid{
    \scriptsize
    \copyright~2026 IEEE. This is the accepted version of an article accepted by \textit{IEEE TPAMI}. The final published version is available at \url{https://doi.org/10.1109/TPAMI.2026.3665753}.
}

\maketitle

\begin{abstract}
Few-step diffusion models enable efficient high-resolution image synthesis but struggle to align with specific downstream objectives due to limitations of existing reinforcement learning (RL) methods in low-step regimes with limited state spaces and suboptimal sample quality. To address this, we propose Stepwise Diffusion Policy Optimization (SDPO), a novel RL framework tailored for few-step diffusion models. SDPO introduces a dual-state trajectory sampling mechanism, tracking both noisy and predicted clean states at each step to provide dense reward feedback and enable low-variance, mixed-step optimization. For further efficiency, we develop a latent similarity-based dense reward prediction strategy to minimize costly dense reward queries. Leveraging these dense rewards, SDPO optimizes a dense reward difference learning objective that enables more frequent and granular policy updates. Additional refinements, including stepwise advantage estimates, temporal importance weighting, and step-shuffled gradient updates, further enhance long-term dependency, low-step priority, and gradient stability. Our experiments demonstrate that SDPO consistently delivers superior reward-aligned results across diverse few-step settings and tasks. Code is available at \url{https://github.com/ZiyiZhang27/sdpo}.
\end{abstract}

\begin{IEEEkeywords}
Diffusion models, text-to-image, few-step diffusion, reinforcement learning, dense rewards.
\end{IEEEkeywords}

\begin{figure*}[t]
    \centering
    \includegraphics[width=\textwidth]{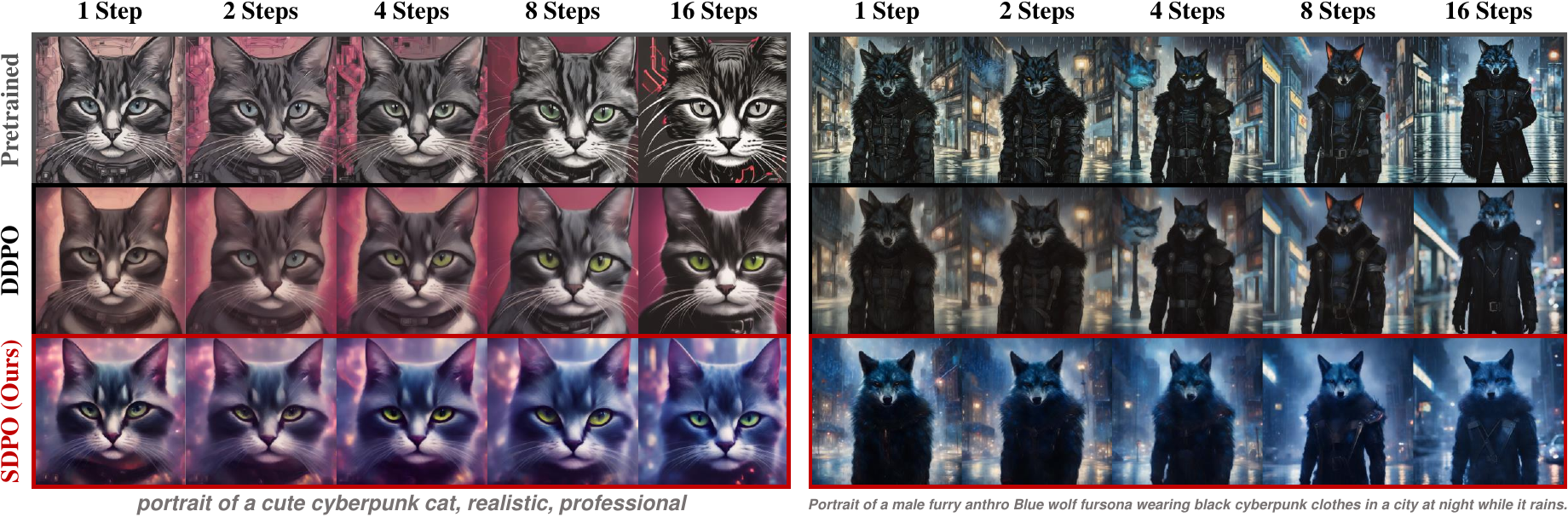}
    \caption{\textbf{Generated images for unseen prompts from:} a pretrained few-step diffusion model (SD-Turbo~\cite{add}), and models finetuned with DDPO~\cite{ddpo} and our SDPO, both using PickScore~\cite{pick} and the same number of training samples. All images are generated using the same random seed (42). Our SDPO consistently delivers high-quality, reward-aligned images across various few-step settings, whereas DDPO falters in generating high-quality few-step samples and yields noticeably blurrier images than even the pretrained model.}
    \label{fig:teaser}
\end{figure*}

\section{Introduction}
\label{sec:1}

\IEEEPARstart{R}{ecent} studies~\cite{salimans2022progressive,meng2023on,add,luo2023diff,cm,lcm,ctm,instaflow,lightning,dmd,dmd2,nguyen2024cvpr} have demonstrated progress in distilling standard text-to-image diffusion models~\cite{d2015,sd,sdxl,sd3} into \textit{few-step} generators, significantly reducing the number of denoising steps needed for high-resolution image synthesis. However, these few-step diffusion models may not inherently align with specific downstream objectives, such as aesthetic quality~\cite{aes} or user preferences~\cite{imagereward,hpsv2,pick,richhf}, which are typically evaluated through reward-based metrics.

\textit{Reinforcement learning (RL) finetuning} \cite{sftpg,dpok,ddpo} offers a promising approach for aligning diffusion models with reward-driven objectives. However, most existing RL methods are primarily designed for standard diffusion models, where state trajectories are sampled with a uniform number of steps (e.g., 20-50 steps). When naively applied to the \textit{few-step} regime (e.g., 1-4 steps), these methods falter in training stability due to the limited state space and suboptimal sample quality of few-step trajectories, leading to insufficient signal diversity for effective reward optimization. On the other hand, incorporating extended trajectories (e.g., 8-50 steps) can enhance output quality and state coverage, but it also leads to overfitting to \textit{sparse}, \textit{final-step} rewards that lack dense guidance for intermediate \textit{low-step} predictions, ultimately impairing few-step performance. While mixed-step optimization broadens step coverage by leveraging sample-reward pairs from trajectories of varying lengths, the resulting high variance across such heterogeneous trajectories destabilizes policy optimization.

\IEEEpubidadjcol

To address these challenges, we first introduce a \textit{dual-state trajectory sampling} mechanism that leverages the strong single-step denoising capability of advanced few-step diffusion models to concurrently track both the \textit{noisy state} and the \textit{predicted clean state} (an intermediate estimate of the final image) at each step. By mapping final outputs from trajectories of varying lengths onto a shared sequence of intermediate clean states, this mechanism enables \textit{dense, stepwise} reward feedback over uniform-step trajectories, facilitating mixed-step optimization with low variance and consistent denoising dynamics across all steps. This not only mitigates overfitting to the final-step outputs but also avoids the high variance inherent in naive mixed-step optimization. To further improve efficiency, we propose a \textit{dense reward prediction} strategy based on \textit{latent similarity}. Instead of querying rewards at every step, we query only at adaptively selected anchor steps and infer the rest via similarity-weighted interpolation, grounded in a smoothness assumption based on Lipschitz continuity.

Leveraging these dense rewards, we formulate a \textit{dense reward difference learning} objective that optimizes stepwise reward differences rather than aggregated trajectory returns, enabling more frequent and granular policy updates compared to trajectory-level methods~\cite{prdp,rebel}. Building on this, we propose \textit{Stepwise Diffusion Policy Optimization} (SDPO), a novel RL finetuning framework tailored for few-step diffusion models. SDPO integrates dual-state sampling to derive dense rewards and incorporates long-term temporal dependencies by computing stepwise advantage estimates, thereby establishing a more robust \textit{stepwise reward difference learning objective}. Additionally, SDPO prioritizes critical low-step optimization through a \textit{temporal importance weighting} strategy and further enhances training stability with \textit{step-shuffled gradient updates}. Together, these innovations empower SDPO to achieve robust and efficient reward optimization, especially in extremely low-step regimes where existing methods tend to falter.

Experimental results show that SDPO excels in reward optimization under extremely low-step regimes, across diverse tasks such as text-to-image generation and multiview image synthesis. Our main contributions are summarized as follows:
\begin{itemize}
\item We introduce a dual-state trajectory sampling mechanism tailored for few-step diffusion models, enabling dense reward feedback and low-variance mixed-step optimization.
\item We develop a latent similarity-based dense reward prediction strategy that minimizes costly reward queries.
\item We formulate a dense reward difference learning objective, allowing few-step diffusion models to effectively optimize dense reward differences at individual steps.
\item We propose the unified SDPO framework, which further enhances dense reward difference learning by stepwise advantage difference learning, temporal importance weighting, and step-shuffled updates for more robust and efficient optimization in extremely low-step regimes.
\end{itemize}

\section{Preliminaries}
\label{sec:2}

\subsection{Diffusion Models}
\label{sec:2.1}

Denoising diffusion probabilistic models~\cite{ddpm} generate high-quality samples by iteratively transforming noise into complex data distributions through a denoising process. In text-to-image applications, these models typically condition each denoising step on a text prompt $\mathbf{c}$ and are operated in two key phases: a forward process that gradually adds noise to an initial sample $\mathbf{x}_0$ at each step $t \in \{1, \dots, T-1\}$, resulting in progressively noisier samples $\mathbf{x}_t$; and a reverse process that iteratively estimates the conditional distribution $p_\theta(\mathbf{x}_{t-1}|\mathbf{x}_t, \mathbf{c})$ at each step to remove noise and reconstruct the original data.

During the reverse process, each denoising step $t$ yields an estimate of the underlying clean sample $\mathbf{x}_0$, often referred to as the \textit{predicted original sample} or \textit{denoised sample}, denoted as $\hat{\mathbf{x}}_0^{t-1}$. For instance, in DDIM~\cite{ddim}, each denoising step $t$ starts by estimating the noise component $\epsilon_\theta(\mathbf{x}_t, t, \mathbf{c})$ from the input sample $\mathbf{x}_t$. This estimated noise is then used to compute the next predicted original sample $\hat{\mathbf{x}}_0^{t-1}$ as:
\begin{equation}
   \hat{\mathbf{x}}_0^{t-1} = \left( \mathbf{x}_t - \sqrt{1 - \bar{\alpha}_t} \cdot \epsilon_\theta(\mathbf{x}_t, t, \mathbf{c}) \right) / \sqrt{\bar{\alpha}_t},
    \label{eq:denoising}
\end{equation}
where $\bar{\alpha}_t$ is the cumulative product of the noise scaling factors $\alpha_t$ up to step $t$. This intermediate prediction of the clean image is progressively refined at each subsequent step.

\subsection{Reinforcement Learning Finetuning for Diffusion Models}
\label{sec:2.2}

\textbf{Denoising as a Markov decision process (MDP).} To perform RL finetuning for diffusion models, existing works~\cite{dpok,ddpo} formulate the denoising process of diffusion models as a $T$-step MDP, which is defined as follows:
\begin{align}
    & s_t \triangleq (\mathbf{c}, \mathbf{x}_{T-t}), \quad a_t \triangleq \mathbf{x}_{T-t-1}, \label{eq:mdp1} \\
    & \pi(a_t | s_t) \triangleq p_\theta(\mathbf{x}_{T-t-1} | \mathbf{x}_{T-t}, \mathbf{c}), \label{eq:mdp2} \\
    & r(s_t, a_t) \triangleq
    \begin{cases}
        R(\mathbf{x}_0, \mathbf{c}), & \text{if}\ t = T-1 \\
        0, & \text{otherwise,}
    \end{cases} \label{eq:mdp3}
\end{align}
where the denoising process $p_\theta$ governed by diffusion model parameters $\theta$ serves as the policy. Starting from an initial noisy state $\mathbf{x}_T$, at each step $t$, the policy treats the denoised sample $\mathbf{x}_{T-t}$ as the current state $s_t$, and the next state $\mathbf{x}_{T-t-1}$ as the action $a_t$. Rewards are defined sparsely, with only the final state $\mathbf{x}_0$ being evaluated via the reward function $R$. The RL objective is thus to maximize the expected reward of final denoising outputs conditioned on prompts $\mathbf{c} \sim p(\mathbf{c})$, i.e.,
\begin{equation}
    \min_{\theta} ~ \mathbb{E}_{p(\mathbf{c})}\mathbb{E}_{p_\theta(\mathbf{x}_0 | \mathbf{c})}\left[ -R(\mathbf{x}_0, \mathbf{c}) \right].
    \label{eq:obj1}
\end{equation}

\textbf{Optimization via policy gradient.} Framing the denoising process as a diffusion policy enables exact computation of log-likelihoods and their gradients w.r.t. the diffusion model parameters $\theta$. Given a full denoising trajectory $\mathbf{x}_{0:T}$, we can specifically compute the log-likelihood $\log{p_{\theta} (\mathbf{x}_{t-1} | \mathbf{x}_t, \mathbf{c})}$ for each denoising step. This allows for the computation of the log-likelihood gradient $\nabla_\theta \log{p_{\theta} (\mathbf{x}_{t-1} | \mathbf{x}_t, \mathbf{c})}$, which captures the sensitivity of each transition probability to changes in the model parameters. By exploiting this property, existing works~\cite{dpok,ddpo} apply policy gradient methods~\cite{reinforce,ppo} to estimate the gradients of Eq.~(\ref{eq:obj1}) as follows:
\begin{equation}
        \mathbb{E}_{p(\mathbf{c})}\mathbb{E}_{p_\theta(\mathbf{x}_{0:T} | \mathbf{c})}\left[ -R(\mathbf{x}_0, \mathbf{c}) \sum_{t=1}^T \nabla_\theta \log{p_{\theta} (\mathbf{x}_{t-1} | \mathbf{x}_t, \mathbf{c})} \right].
    \label{eq:obj2}
\end{equation}

\textbf{Reward difference learning.} Inspired by direct preference optimization (DPO)~\cite{dpo} that converts the RL objective for language models into a supervised classification objective, recent approaches~\cite{prdp,rebel} further derive a mean squared error-based objective for diffusion models, where the goal is to match the differences in log-likelihood ratios with the differences in rewards across paired denoising trajectories:
\begin{equation}
    \begin{aligned}
        \mathbb{E}_{\mathbf{x}^a, \mathbf{x}^b, \mathbf{c}} \Bigg[ \sum_{t=1}^T
            & \bigg( \frac{1}{\eta} \left( \log \frac{p_\theta(\mathbf{x}^a_t | \mathbf{c})}{p_{\theta'}(\mathbf{x}^a_t | \mathbf{c})} 
            - \log \frac{p_\theta(\mathbf{x}^b_t | \mathbf{c})}{p_{\theta'}(\mathbf{x}^b_t | \mathbf{c})} \right) \bigg. \Bigg. \\
            &\Bigg. \bigg. - \left( R(\mathbf{x}^a_0, \mathbf{c}) - R(\mathbf{x}^b_0, \mathbf{c}) \right) \bigg)^2
        \Bigg].
    \end{aligned}
    \label{eq:obj3}
\end{equation}
Here, $\mathbf{x}^a_t$ and $\mathbf{x}^b_t$ represent paired samples from the denoising step $t$ of two independent trajectories, both generated using the same text prompt $\mathbf{c}$. These paired samples are used to compute the differences in their corresponding rewards, i.e., $R(\mathbf{x}^a_0, \mathbf{c}) - R(\mathbf{x}^b_0, \mathbf{c})$, as well as the differences in their log-likelihood ratios. In particular, $\log \frac{p_\theta(\mathbf{x}^a_t | \mathbf{c})}{p_{\theta'}(\mathbf{x}^a_t | \mathbf{c})}$ denotes the log-likelihood ratio of the sample $\mathbf{x}^a_t$ under the current model parameters $\theta$ compared to the previous or reference model parameters $\theta'$, helping to quantify the change in the model during online optimization. Furthermore, the \textit{log-ratio scale factor} $\eta$ balances the contribution of log-ratio differences relative to reward differences in the objective. Empirically, both Deng et al.~\cite{prdp} and Gao et al.~\cite{rebel} demonstrate that this reward difference-based approach results in more stable reward optimization for diffusion models, compared to vanilla policy gradient methods~\cite{dpok,ddpo}.

\section{Method}
\label{sec:3}

\subsection{Limitations of Existing RL Finetuning for Few-Step Diffusion Models}
\label{sec:3.1}

Existing RL finetuning methods for diffusion models typically rely on a fixed, uniform number of denoising steps per trajectory (e.g., 50 steps). While effective for standard diffusion models, this uniform-step paradigm introduces critical challenges in RL Finetuning for few-step diffusion models. In the target few-step regime (e.g., 1 to 4 steps), short sampling trajectories result in limited state spaces and inherently suboptimal output quality, yielding insufficient signal diversity for reliable policy updates. As a result, naively applying uniform-step RL finetuning to few-step diffusion models leads to unstable training dynamics and poor sample efficiency (as evidenced by the reward curves in Fig.~\ref{fig:instab}).

On the other hand, using extended trajectories (e.g., 8 to 50 steps) can generate higher-quality samples and expand the state space. However, this approach has its own drawbacks. Without dense, step-by-step reward feedback for intermediate low-step samples, the model tends to overfit to the longer trajectories, undermining its few-step inference ability (see the reward curves for low-step samples in Fig.~\ref{fig:sample}). However, deriving dense rewards for intermediate steps is not straightforward, since most existing reward functions are designed to evaluate final clean images rather than the noisy intermediate states typical of denoising processes. One potential solution is to mix trajectories of different lengths, providing rewards at various sampling steps. While this can offer more generalization, the inconsistent dynamics across trajectories introduce significant variance, which destabilizes policy optimization and undermines training consistency (as evidenced by the reward curves in the right plot of Fig.~\ref{fig:instab}).

\begin{figure}[t]
    \centering
    \includegraphics[width=\columnwidth]{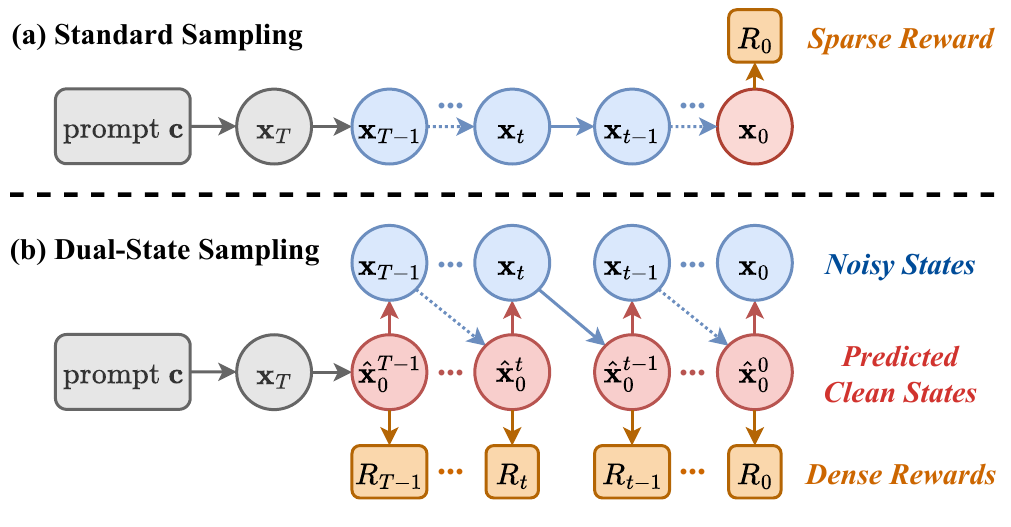}
    \caption{\textbf{Dual-state sampling vs.\ standard sampling.} Unlike the standard sampling process of diffusion models, our dual-state sampling approach maps final outputs from trajectories of varying lengths onto a shared sequence of intermediate clean states $\{\hat{\mathbf{x}}_0^t\}_{t=0}^{T-1}$, enabling dense reward feedback over a mixed-step trajectory with low variance and consistent denoising dynamics.}
    \label{fig:dual}
\end{figure}

\subsection{Mixed-Step Trajectories with Consistent Dynamics}
\label{sec:3.2}

To address these challenges, we propose a \textit{dual-state sampling} mechanism that enables dense, stepwise reward feedback over uniform-step trajectories, facilitating mixed-step trajectory optimization with low variance and consistent denoising dynamics across all steps. Crucially, while naive mixed-step denoising trajectories suffer from high variance due to inconsistent denoising dynamics, this dual-state sampling mechanism maps final outputs from trajectories of varying lengths onto a \textit{shared, consistent} sequence of predicted clean states $\{\hat{\mathbf{x}}_0^t\}_{t=0}^{T-1}$ across all steps, effectively reducing this variance.

\textbf{Dual-state sampling.}
As discussed in Section~\ref{sec:2.1}, the \textit{predicted original sample} $\hat{\mathbf{x}}_0^{t-1}$ serves as an intermediate estimate of the \textit{final, noise-free} data at the corresponding denoising step $t$. Due to the strong single-step denoising capacity distilled into few-step diffusion models, this intermediate estimate remains highly accurate even during early denoising steps, allowing $\hat{\mathbf{x}}_0^{t-1}$ to act as a reliable \textit{surrogate} for the final output of a full $t$-step denoising process. Building on this insight, our dual-state sampling mechanism concurrently tracks two states throughout the trajectory: the \textit{noisy} state $\mathbf{x}_{t-1}$ and the predicted \textit{clean} state $\hat{\mathbf{x}}_0^{t-1}$. As illustrated in Fig.~\ref{fig:dual}~(b), at each sampling step $t$, we first denoise the previous noisy state $\mathbf{x}_t$ to compute the predicted clean state $\hat{\mathbf{x}}_0^{t-1}$ via Eq.~(\ref{eq:denoising}). The next noisy state $\mathbf{x}_{t-1}$ is then derived by combining $\hat{\mathbf{x}}_0^{t-1}$ with the directional noise component:
\begin{equation}
    \mathbf{x}_{t-1} = \sqrt{\bar{\alpha}_{t-1}} \hat{\mathbf{x}}_0^{t-1} + \sqrt{1 - \bar{\alpha}_{t-1}} \cdot \epsilon_\theta(\mathbf{x}_t, t, \mathbf{c}).
\label{eq:dual}
\end{equation}
This approach preserves the trajectory's stochastic dynamics while maintaining a direct link to intermediate clean estimates, laying the foundation for the dense reward feedback.

\textbf{Dense reward feedback via $\hat{\mathbf{x}}_0^t$.}
The dual-state sampling mechanism enables us to evaluate and assign dense, step-level rewards at each intermediate denoising step $t$ by leveraging the predicted clean estimates $\hat{\mathbf{x}}_0^{t-1}$ within the dual-state trajectory, rather than relying solely on sparse rewards from the final-step outputs. Accordingly, we redefine the trajectory reward in the denoising MDP (Eq.~(\ref{eq:mdp3})) by applying the predefined reward function $R$ to each intermediate clean state $\hat{\mathbf{x}}_0^{T-t-1}$:
\begin{equation}
    r(s_t, a_t) \triangleq R(\hat{\mathbf{x}}_0^{T-t-1}, \mathbf{c}).
    \label{eq:mdp4}
\end{equation}
This reformulation ensures that each denoising step contributes directly to policy optimization, guided by the quality of its corresponding clean estimate. By decoupling trajectory length from reward structure, our approach not only prevents overfitting to the high-step, final outputs of uniform-length trajectories, but also eliminates the high variance inherent in naive mixed-step trajectories. This yields a more balanced mixed-step optimization paradigm, where the progressive refinement of $\hat{\mathbf{x}}_0^t$ over steps drives consistent and stable optimization.

\begin{figure*}[t]
    \centering
    \includegraphics[width=\textwidth]{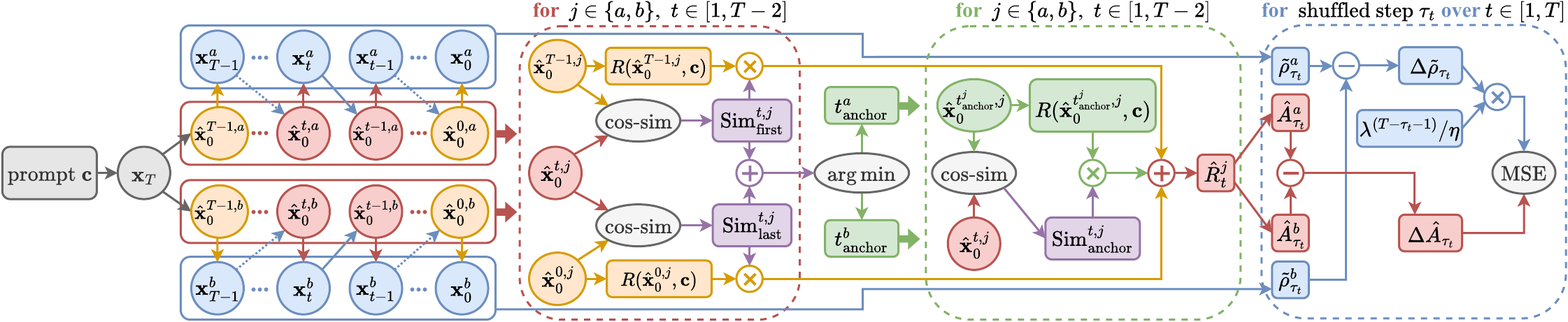}
    \caption{\textbf{SDPO framework.} SDPO first samples a pair of \textit{dual-state trajectories} $\{\mathbf{x}_t^{a}, \hat{\mathbf{x}}_0^{t,a}\}_{t=0}^{T-1}$ and $\{\mathbf{x}_0^{t,b}, \hat{\mathbf{x}}_0^{t,b}\}_{t=0}^{T-1}$ using a shared \textit{prompt} $c$ and \textit{initial noise} $\mathbf{x}_T$. It then queries the \textit{reward function} $R$ for clean states at the \textit{first}, \textit{final}, and \textit{anchor} ($t_{\text{anchor}}$) steps and predicts \textit{dense rewards} $\hat{R}_t$ for other steps via latent similarity, ultimately yielding the \textit{stepwise advantage estimate} $\hat{A}_t$. Finally, at each \textit{shuffled step} $\tau_t$, the MSE loss between the \textit{advantage difference} $\Delta\hat{A}_{\tau_t}$ and the \textit{log-ratio difference} $\Delta\tilde{\rho}_{\tau_t}$ (weighted by $\lambda^{(T-\tau_t-1)}/\eta$) is computed.}
    \label{fig:sdpo}
\end{figure*}

\subsection{Dense Reward Difference Learning for Few-Step Diffusion Models}
\label{sec:3.3}

Although our dual-state sampling enables dense rewards from intermediate clean states, directly decoding the latent-space states to image space for reward queries remains computationally prohibitive. To address this, we propose an efficient dense reward prediction strategy based on latent similarity, reducing reward queries while maintaining robust and granular guidance throughout the denoising process.

\textbf{Dense reward prediction via latent similarity.}
Rather than querying rewards at every denoising step, we limit reward queries to three per trajectory: two fixed queries at the first and last steps, and one query at an adaptively selected intermediate step, referred to as the \textit{anchor step} ($t_{\text{anchor}}$). To determine the anchor step, we first calculate the cosine similarity between the latent representations of each predicted clean state $\hat{\mathbf{x}}_0^t$ and those at the first and last steps, respectively:
\begin{equation}
    \text{Sim}_{\text{first}}^t = \frac{\langle \hat{\mathbf{x}}_0^t, \hat{\mathbf{x}}_0^{T-1} \rangle}{\|\hat{\mathbf{x}}_0^t\| \|\hat{\mathbf{x}}_0^{T-1}\|}, \quad \text{Sim}_{\text{last}}^t = \frac{\langle \hat{\mathbf{x}}_0^t, \hat{\mathbf{x}}_0^0 \rangle}{\|\hat{\mathbf{x}}_0^t\| \|\hat{\mathbf{x}}_0^0\|}.
\end{equation}
The anchor step is then identified as the step that minimizes the sum of cosine similarities to the first and last steps:
\begin{equation}
    t_{\text{anchor}} = \underset{t \in \{1, \dots, T-2\}}{\arg\min} \left( \text{Sim}_{\text{first}}^t + \text{Sim}_{\text{last}}^t \right),
    \label{eq:anchor}
\end{equation}
which effectively selects the step that is most distinct from both endpoints in the latent space. This adaptive selection strategy is grounded in information theory and optimal sampling principles~\cite{shewry1987maximum,belhadji2020kernel}, aiming to maximize the total information gained from the limited reward queries by minimizing redundancy among them. By choosing an anchor step that is minimally correlated with two extremes of the trajectory, we ensure that the three queried steps provide maximal insight into the dense reward landscape, thereby enhancing the reliability of subsequent reward predictions for unqueried steps.

With rewards queried at these three steps, we predict the dense reward $\hat{R}_t$ for each intermediate step $t$ by performing a similarity-weighted interpolation of the queried rewards:
\begin{equation}
    \hat{R}_t = \frac{
        R_{T-1} \cdot \text{Sim}_{\text{first}}^t + 
        R_{t_{\text{anchor}}} \cdot \text{Sim}_{\text{anchor}}^t + 
        R_{0} \cdot \text{Sim}_{\text{last}}^t
    }{
        \text{Sim}_{\text{first}}^t + 
        \text{Sim}_{\text{anchor}}^t + 
        \text{Sim}_{\text{last}}^t
    }
    \label{eq:dense}
\end{equation}
where $R_{T-1}$, $R_{t_{\text{anchor}}}$, and $R_0$ denote the rewards queried at the first, anchor, and last steps, respectively, and $\text{Sim}_{\text{anchor}}^t$ is the cosine similarity between $\hat{\mathbf{x}}_0^t$ and $\hat{\mathbf{x}}_0^{t_{\text{anchor}}}$. This interpolation leverages the latent similarities to effectively estimate rewards for unqueried steps, providing dense feedback while minimizing the number of costly reward function evaluations.

\textbf{Smoothness assumption via Lipschitz continuity.} Our dense reward prediction strategy is grounded in a smoothness assumption: the composition of the reward function $R$ with the latent encoder $\boldsymbol{\phi}$ satisfies a Lipschitz condition. Formally, there exists a constant $K_R > 0$ such that
\begin{equation}
    \bigl|R\bigl(\hat{\mathbf{x}}_0^i,\mathbf{c}\bigr) - R\bigl(\hat{\mathbf{x}}_0^j,\mathbf{c}\bigr)\bigr|
    \, \leq \, K_R\,\bigl\|\boldsymbol{\phi}(\hat{\mathbf{x}}_0^i) - \boldsymbol{\phi}(\hat{\mathbf{x}}_0^j)\bigr\|.
\end{equation}
This Lipschitz property ensures that small perturbations in the latent representation lead to bounded changes in reward values, justifying our use of latent similarity as a reliable proxy for reward interpolation. This assumption holds for many practical reward functions, particularly those based on perceptual quality or semantic alignment, which typically exhibit smooth variations with respect to image content.

\textbf{Dense reward difference learning.}
Leveraging dense rewards derived from two independent dual-state trajectories, i.e., $\hat{R}_t^a$ and $\hat{R}_t^b$, we formulate a dense reward difference learning objective that aligns the difference in \textit{dense rewards} with the difference in \textit{log-likelihood ratios} at each step $t$. Specifically, we define the log-likelihood ratio for each trajectory as $\rho^j_t(\theta) := \log (p_\theta(\mathbf{x}^j_t | \mathbf{c})/p_{\theta'}(\mathbf{x}^j_t | \mathbf{c}))$, where $j \in \{a, b\}$. The objective is then formulated as:
\begin{equation}
    \mathbb{E}_{\mathbf{x}_t^a, \mathbf{x}_t^b, \mathbf{c}} \Bigg[
        \left( \frac{1}{\eta} \left( \rho^a_t(\theta) - \rho^b_t(\theta) \right) - \left( \hat{R}_t^a - \hat{R}_t^b \right) \right)^2
    \Bigg],
    \label{eq:obj4}
\end{equation}
where $\theta'$ refers to the previous model parameters. By eliminating the need to accumulate full trajectory data, this stepwise objective enables more frequent and granular updates compared to trajectory-level objectives like Eq.~(\ref{eq:obj3}).

\subsection{Stepwise Diffusion Policy Optimization}
\label{sec:3.4}

Building on the dual-state sampling mechanism from Section~\ref{sec:3.2} and the dense reward difference learning objective from Section~\ref{sec:3.3}, we propose \textit{Stepwise Diffusion Policy Optimization} (SDPO), a novel RL finetuning framework designed to effectively optimize few-step diffusion models through stepwise, dense reward feedback. Unlike existing methods that optimize final outputs from full trajectories monolithically, SDPO exploits the stepwise granularity of intermediate clean states and their dense rewards to tackle the unique challenges inherent in optimizing few-step denoising outputs, while ensuring stable and efficient policy updates.

\textbf{Dual-state trajectory integration.} First, SDPO unifies dual-state trajectory sampling from Section~\ref{sec:3.2} and dense reward difference learning from Section~\ref{sec:3.3} within a single Markovian framework. Leveraging the dual-state sampling mechanism, SDPO simultaneously tracks the noisy state $\mathbf{x}_t$ (for policy rollouts) and its corresponding predicted clean state $\hat{\mathbf{x}}_0^t$ (for dense reward prediction via Eq.~(\ref{eq:dense})) at each step $t$. Incorporated with the dense reward difference learning objective, SDPO generates two independent dual-state trajectories $\{\mathbf{x}_t^{a}, \hat{\mathbf{x}}_0^{t,a}\}_{t=0}^{T-1}$ and $\{\mathbf{x}_t^{b}, \hat{\mathbf{x}}_0^{t,b}\}_{t=0}^{T-1}$ under the same initial noise $\mathbf{x}_T$ and prompt $\mathbf{c} \sim p(\mathbf{c})$. In contrast to standard reward difference learning methods that use independently initialized noise for each trajectory, SDPO shares initial noise between each trajectory pair to reduce variance in log-likelihoods, thereby leading to more stable and efficient policy updates.

\textbf{Stepwise advantage difference learning.} While Eq.~(\ref{eq:obj4}) enables granular, step-by-step policy updates, it overlooks the temporal dependencies inherent in the denoising process, where early denoising steps shape later outcomes. To address this, SDPO incorporates long-term feedback from future steps into the current step's evaluation by computing a \textit{discounted return} $\hat{G}_t^j$ for each trajectory $j \in \{a, b\}$ at step $t$, aggregating dense rewards from future steps with a discount factor $\gamma \in \left(0, 1\right]$, i.e., $\hat{G}_t^j = \sum_{k=0}^{t} \gamma^{k} \hat{R}_{t-k}^j$. To reduce variance across trajectories while preserving inter-step relationships, SDPO performs a novel \textit{per-step-prompt normalization} mechanism on $\hat{G}_t^j$, maintaining running estimates of the mean $\mu_t$ and standard deviation $\sigma_t$ for each step-prompt pair. This yields \textit{stepwise advantage estimates} $\hat{A}_t^j$, which quantify relative improvement over baseline performance. By substituting rewards in Eq.~(\ref{eq:obj4}) with these stepwise advantage estimates, SDPO induces a \textit{stepwise advantage difference learning} objective:
\begin{equation}
    \mathbb{E}_{\mathbf{x}_t^a, \mathbf{x}_t^b, \mathbf{c}} \Bigg[
        \left( \frac{1}{\eta} \left( \rho^a_t(\theta) - \rho^b_t(\theta) \right) - \left( \hat{A}_t^a - \hat{A}_t^b \right) \right)^2
    \Bigg].
\end{equation}

\begin{algorithm}[tb]
    \caption{Stepwise Diffusion Policy Optimization}
    \label{alg:SDPO}
    \begin{algorithmic}[1]
        \STATE \textbf{Input:} Pretrained diffusion model parameters $\theta$, prompt distribution $p(\mathbf{c})$, number of training epochs $E$, number of denoising timesteps per trajectory $T$, batch size $B$
        \FOR{epoch $e = 1, \dots, E$}
            \STATE Sample a batch of text prompts $\{ \mathbf{c}_i \sim p(\mathbf{c}) \}_{i=1}^{B}$
            \STATE \textit{Phase 1: Trajectory Sampling (no gradient)}
            \FOR{mini-batch $i = 1, \dots, B$}
                \STATE Generate dual-state trajectory $\{\mathbf{x}_t^{a,i}, \hat{\mathbf{x}}_0^{t,a,i}\}_{t=0}^{T-1}$
                \STATE Generate dual-state trajectory $\{\mathbf{x}_t^{b,i}, \hat{\mathbf{x}}_0^{t,b,i}\}_{t=0}^{T-1}$
                \STATE Compute dense rewards $\{\hat{R}_t^{a,i}, \hat{R}_t^{b,i}\}_{t=0}^{T-1}$
                \STATE Compute discounted returns $\{\hat{G}_t^{a,i}, \hat{G}_t^{b,i}\}_{t=0}^{T-1}$
                \STATE Generate shuffled step indices $\{\tau_t^i\}_{t=0}^{T-1}$
            \ENDFOR
            \STATE Compute advantage estimates $\{\hat{A}_t^{a,i}, \hat{A}_t^{b,i}\}_{t=0}^{T-1}$
            
            \STATE \textit{Phase 2: Policy Update (with gradient)}
            \FOR{timestep $t = 1, \dots, T$}
                \STATE Re-generate next states at shuffled steps $\{\tau_t^i\}_{i=1}^{B}$
                \STATE Compute log-likelihoods at shuffled steps $\{\tau_t^i\}_{i=1}^{B}$
                \STATE Update $\theta$ by minimizing $\mathcal{L}_t(\theta)$ in Eq.(\ref{eq:update})
            \ENDFOR
        \ENDFOR
    \end{algorithmic}
\end{algorithm}

\textbf{Temporal importance weighting.} To prioritize optimization for low-step samples while still utilizing later-step samples for broader state space exploration, we introduce a \textit{temporal importance weighting} strategy in SDPO's dense reward difference learning objective. This is achieved by applying an exponentially decaying weight, controlled by the \textit{decay rate} $\lambda$, to the log-likelihood ratios. Specifically, the weight for a sample at step $t$ is defined as $\lambda^{(T-t-1)}$, decreasing from the first step ($t=T-1$) to the last step ($t=0$). In practice, this means that the log-ratio difference, $\Delta\rho_t(\theta) := \rho^a_t(\theta) - \rho^b_t(\theta)$, is scaled by $\lambda^{(T-t-1)}/\eta$ and matched against the corresponding advantage difference, $\Delta\hat{A}_t := \hat{A}_t^a - \hat{A}_t^b$. Thus, the weighted objective function for each step $t$ can be formulated as:
\begin{equation}
    l_\theta = \mathbb{E}_{\mathbf{x}_t^a, \mathbf{x}_t^b, \mathbf{c}} \left[ \left( \Delta\rho_t(\theta) \cdotp \lambda^{(T-t-1)}/\eta - \Delta\hat{A}_t \right)^2 \right].
    \label{eq:obj5}
\end{equation}
Additionally, to prevent large, unstable policy updates, we clip the log-likelihood ratios at a threshold $\epsilon$, defining $\tilde{\rho}^j_t(\theta) := \text{clip}(\rho^j_t(\theta), -\epsilon, \epsilon)$, yielding the clipped objective:
\begin{equation}
    \tilde{l}_\theta = \mathbb{E}_{\mathbf{x}_t^a, \mathbf{x}_t^b, \mathbf{c}} \left[ \left( \Delta\tilde{\rho}_t(\theta) \cdotp \lambda^{(T-t-1)}/\eta - \Delta\hat{A}_t \right)^2 \right].
    \label{eq:obj6}
\end{equation}

\textbf{Step-shuffled gradient updates.} Leveraging the stepwise granularity of our advantage estimates, SDPO performs gradient updates individually for each step rather than aggregating over entire trajectories. However, sequential stepwise updates risk overfitting to the fixed step order of trajectories. To address this, we introduce step-shuffled gradient updates, where the step order within each mini-batch of trajectories is independently shuffled. As illustrated in Algorithm~\ref{alg:SDPO}, SDPO conducts $T$ gradient updates per training epoch, with $T$ being the number of steps per trajectory. For the $t$-th update, gradient estimates are computed by averaging over $B$ mini-batches of stepwise samples ($\mathbf{x}^{a,i}_{\tau_t^i}, \mathbf{x}^{b,i}_{\tau_t^i}, \mathbf{c}_i$), where $\tau_t^i$ denotes the shuffled step index for the $i$-th mini-batch. Accordingly, the empirical training objective of SDPO is defined as:
\begin{equation}
    \mathcal{L}_t(\theta) \leftarrow \frac{1}{B} \sum_{i=1}^{B} \max \left(l_\theta(\mathbf{x}^{a,i}_{\tau_t^i}, \mathbf{x}^{b,i}_{\tau_t^i}, \mathbf{c}_i), \tilde{l}_\theta(\mathbf{x}^{a,i}_{\tau_t^i}, \mathbf{x}^{b,i}_{\tau_t^i}, \mathbf{c}_i) \right),
    \label{eq:update}
\end{equation}
where the maximization between the objectives $l_\theta$ and $\tilde{l}_\theta$ serves to minimize an upper bound on the policy updates. The overall framework of SDPO is illustrated in Fig.~\ref{fig:sdpo}.

\begin{figure*}[t]
    \centering
    \includegraphics[width=0.245\textwidth]{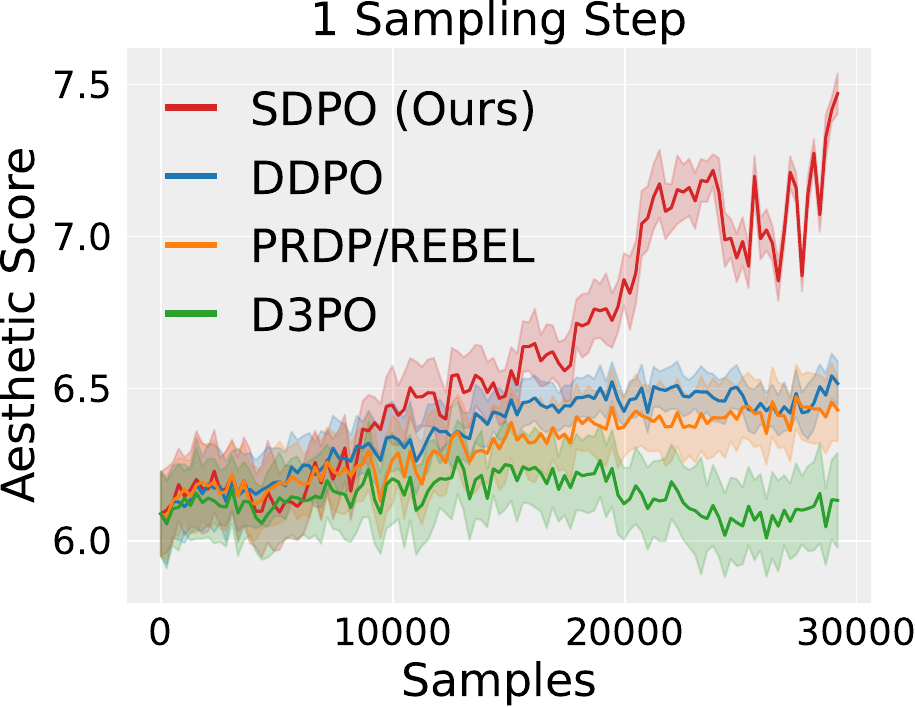}
    \includegraphics[width=0.245\textwidth]{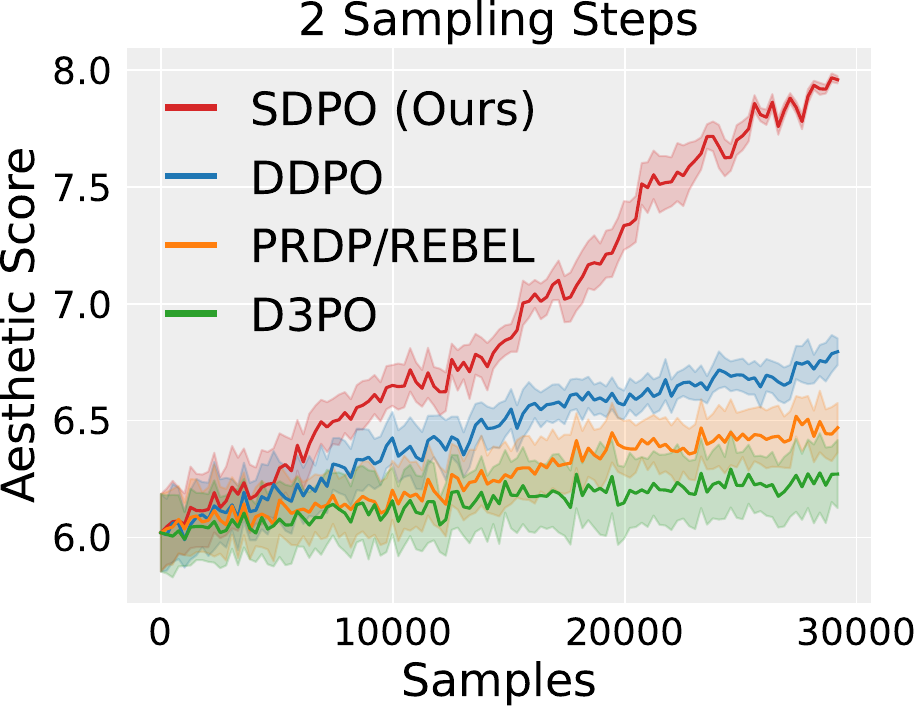}
    \includegraphics[width=0.245\textwidth]{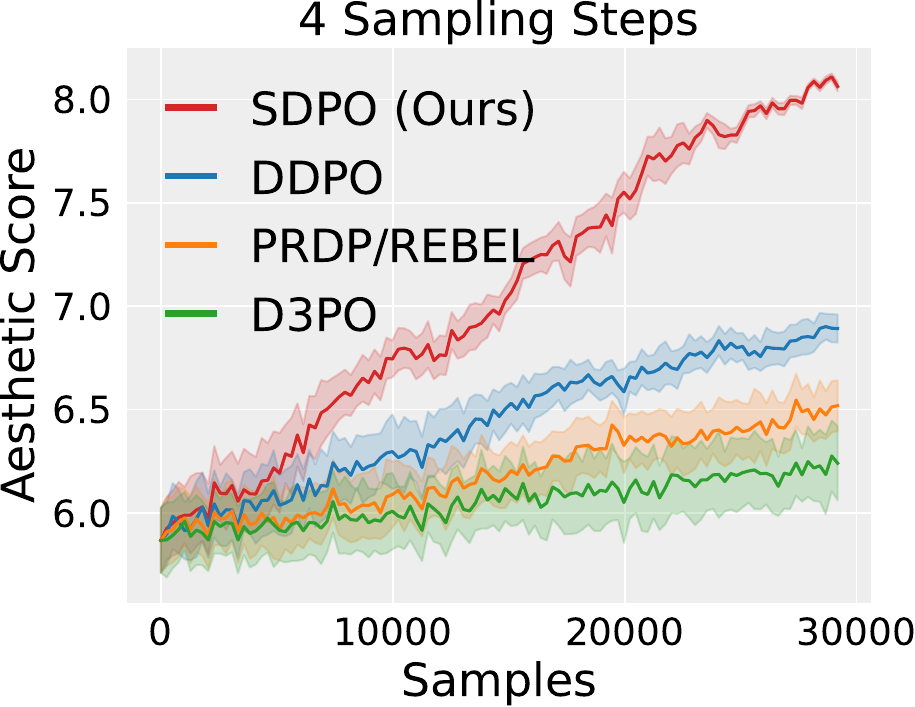}
    \includegraphics[width=0.245\textwidth]{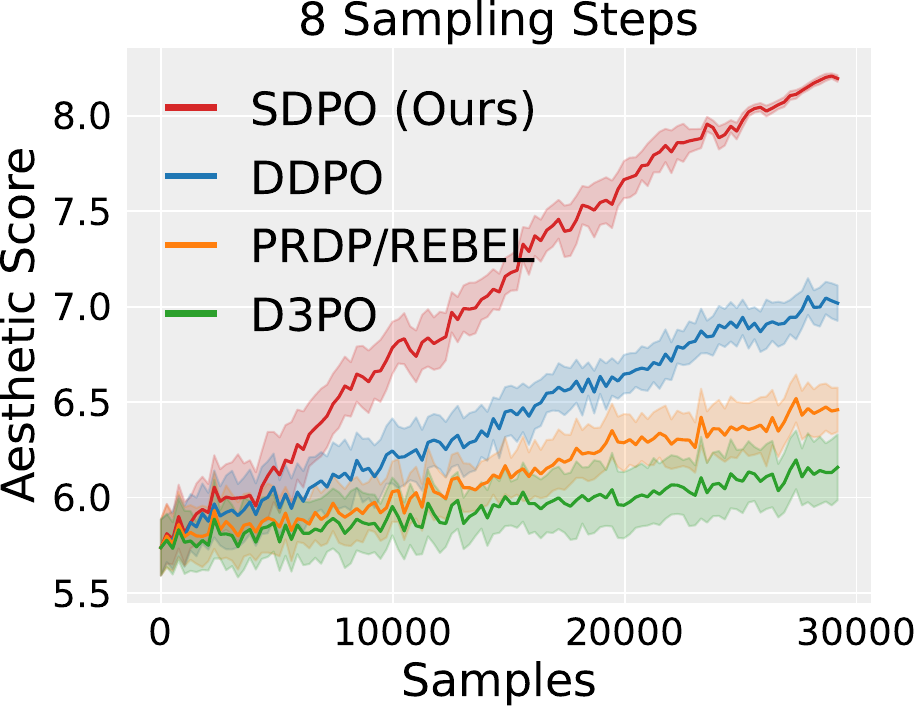}
    \caption{\textbf{Reward curves for low-step samples}, where reward scores are evaluated and averaged over 1-, 2-, 4-, and 8-step samples. The horizontal axis shows the cumulative number of training samples, equivalent to the number of training iterations multiplied by the batch size, which is consistent across all methods.}
    \label{fig:sample}
\end{figure*}

\section{Experiments}
\label{sec:4}

\subsection{Main Experimental Setup}
\label{sec:4.1}
In our main experiments (Sections~\ref{sec:4.2} and \ref{sec:4.3}), we use SD-Turbo~\cite{add}, a few-step text-to-image model distilled from Stable Diffusion v2.1~\cite{sd} through adversarial diffusion distillation~\cite{add}, as the base model. For efficient finetuning, we apply Low-Rank Adaptation (LoRA)~\cite{lora}, which enables updating only the LoRA weights while freezing the pretrained model parameters. Following DDPO~\cite{ddpo}, we generate denoising trajectories during training using the DDIM sampler~\cite{ddim} and a prompt set of 45 animal names. Since SD-Turbo does not support classifier-free guidance or negative prompts, we disable them by setting the guidance scale to $1.0$. We implement four commonly used reward functions: Aesthetic Score~\cite{aes}, ImageReward~\cite{imagereward}, HPSv2~\cite{hpsv2}, and PickScore~\cite{pick}.

\begin{table*}[t]
    \caption{\textbf{Generalization to unseen prompts.} We report reward scores for 1-, 2-, 4-, and 8-step samples generated from unseen prompts by both the non-finetuned base model (labeled as SD-Turbo) and the finetuned models of different RL finetuning methods.}
    \label{tab:unseen}
    \centering
    \resizebox{\textwidth}{!}{
    \begin{tabular}{lcccccccccccccccccccccc}
        \toprule
        \multirow{2}{*}{\vspace{-1mm}\textbf{Method}} & \multicolumn{4}{c}{\textbf{ImageReward}} & \multicolumn{4}{c}{\textbf{Aesthetic Score}} & \multicolumn{4}{c}{\textbf{HPSv2}} & \multicolumn{4}{c}{\textbf{PickScore}}\\
        \cmidrule(lr){2-5} \cmidrule(lr){6-9} \cmidrule(lr){10-13} \cmidrule(lr){14-17}
        & \textit{\textbf{1s}} & \textit{\textbf{2s}} & \textit{\textbf{4s}} & \textit{\textbf{8s}} & \textit{\textbf{1s}} & \textit{\textbf{2s}} & \textit{\textbf{4s}} & \textit{\textbf{8s}} & \textit{\textbf{1s}} & \textit{\textbf{2s}} & \textit{\textbf{4s}} & \textit{\textbf{8s}} & \textit{\textbf{1s}} & \textit{\textbf{2s}} & \textit{\textbf{4s}} & \textit{\textbf{8s}} \\
        \midrule
        SD-Turbo~\cite{add}  & 1.126 & 1.101 & 1.038 & 0.976 & 6.064 & 5.955 & 5.842 & 5.745 & 27.85 & 28.22 & 28.41 & 28.26 & 22.69 & 22.88 & 22.91 & 22.68 \\
        D3PO~\cite{d3po}     & 1.292 & 1.211 & 1.176 & 1.210 & 6.292 & 6.208 & 6.159 & 6.116 & 28.06 & 28.51 & 28.77 & 28.78 & 22.64 & 22.95 & 23.16 & 23.00 \\
        DDPO~\cite{ddpo}     & 1.380 & 1.252 & 1.378 & 1.387 & 6.644 & 6.584 & 6.595 & 6.720 & 28.54 & 28.75 & 28.99 & 29.05 & 22.65 & 22.94 & 23.04 & 23.29 \\
        REBEL~\cite{rebel}   & 1.280 & 1.305 & 1.289 & 1.294 & 6.509 & 6.471 & 6.472 & 6.418 & 28.28 & 28.58 & 28.87 & 28.97 & 22.61 & 22.86 & 23.01 & 23.19 \\
        \rowcolor{blue!10}
        SDPO (Ours)          & \textbf{1.719} & \textbf{1.559} & \textbf{1.545} & \textbf{1.601} & \textbf{7.062} & \textbf{7.516} & \textbf{7.867} & \textbf{8.163} & \textbf{28.77} & \textbf{29.14} & \textbf{29.35} & \textbf{29.36} & \textbf{23.02} & \textbf{23.22} & \textbf{23.60} & \textbf{24.05} \\
        \bottomrule
    \end{tabular}
    }
\end{table*}

\subsection{Algorithm Comparisons}
\label{sec:4.2}
We compare our SDPO against established reward finetuning methods for diffusion models, selecting representatives from three distinct categories. For policy gradient methods, we adopt DDPO~\cite{ddpo} due to its proven effectiveness in this domain. For reward difference-based approaches, we initially targeted PRDP~\cite{prdp}, but its lack of open-source availability led us to implement REBEL~\cite{rebel} instead; however, encountering mode collapse with REBEL on SD-Turbo prompted us to introduce clipping from PRDP, resulting in an interpolated 'PRDP/REBEL' method that mitigates this issue. For preference-based methods, such as Diffusion-DPO~\cite{wallace2024diffusion} and KTO~\cite{li2024aligning}, we note that they typically prioritize learning from data-driven preferences rather than explicit reward functions, distinguishing their objectives from ours. Nevertheless, we adopt D3PO~\cite{d3po}, another state-of-the-art DPO-style method for diffusion models, which leverages preference relationships from relative rewards, enabling reward-based finetuning aligned with our settings. To ensure a fair comparison, we standardize all shared hyperparameters across algorithms (implementation details are presented in the appendix).

\textbf{Sample efficiency in optimizing low-step samples.} Sample efficiency measures how effectively algorithms leverage samples to enhance reward outcomes. We evaluate this by generating samples from intermediate model checkpoints using various step configurations and assessing their reward scores. Fig.~\ref{fig:sample} displays Aesthetic Score reward curves for images generated with 1, 2, 4, and 8 sampling steps, while additional reward curves are provided in the appendix. In these plots, the x-axis specifies the cumulative number of samples used for training, and the y-axis indicates the average reward of low-step samples. The steeper reward curves of SDPO suggest that it is able to achieve higher rewards with fewer samples compared to other algorithms, demonstrating its superior sample efficiency in optimizing low-step samples.

\textbf{Generalization to unseen prompts.} We further evaluate the generalization capability of our SDPO on unseen, complex text prompts from \cite{dpok}, specifying attributes like color (\textit{A green colored rabbit}), count (\textit{Four wolves in the park}), composition (\textit{A cat and a dog}), and location (\textit{A dog on the moon}). Accordingly, Table~\ref{tab:unseen} presents reward scores for 1-, 2-, 4-, and 8-step samples generated from these unseen prompts by both the non-finetuned SD-Turbo and various finetuned models. Notably, SDPO consistently outperforms all baselines across different reward functions and step settings. Qualitative results on unseen prompts are provided in Fig.~\ref{fig:teaser} and the appendix.

\begin{figure*}[t]
    \centering
    \includegraphics[width=0.33\textwidth]{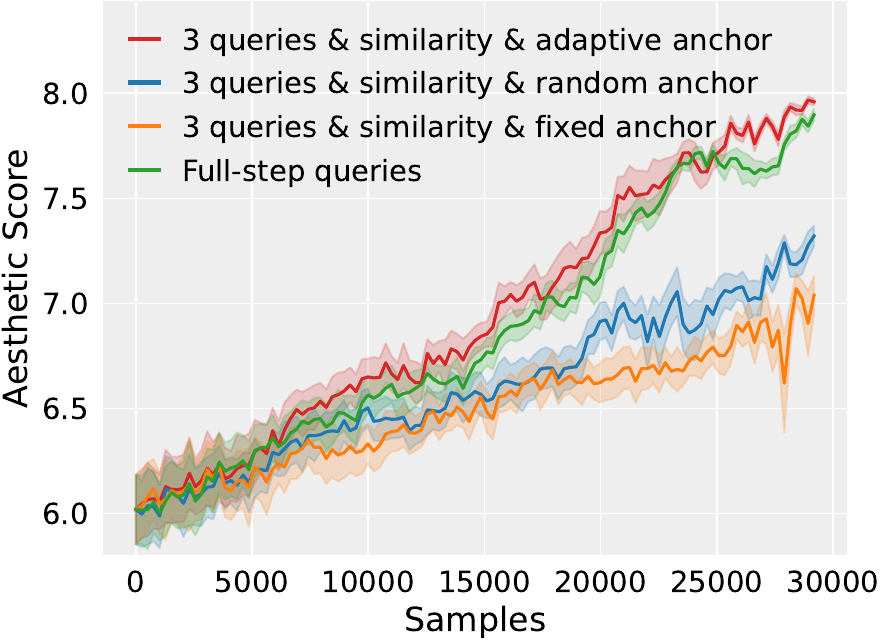}
    \includegraphics[width=0.32\textwidth]{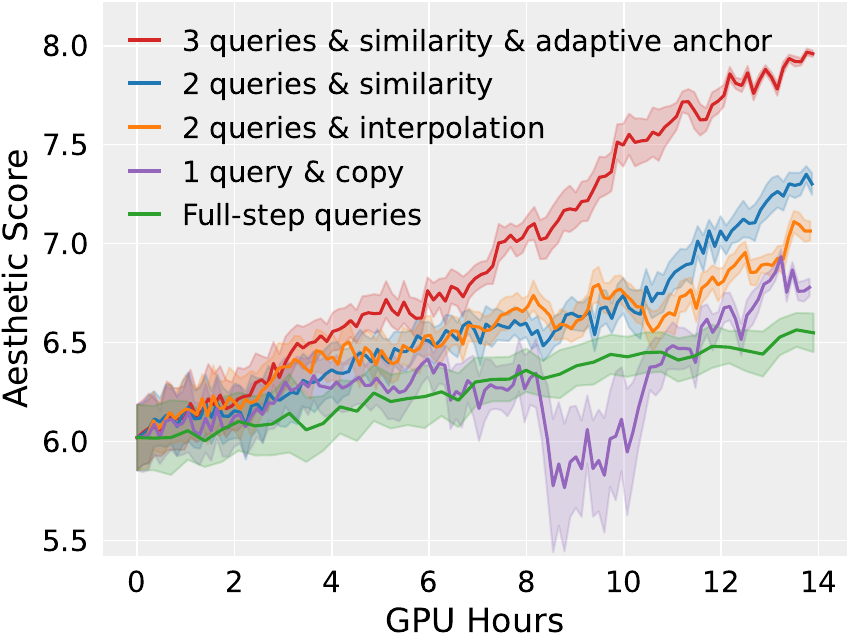}
    \hfill
    \includegraphics[width=0.33\textwidth]{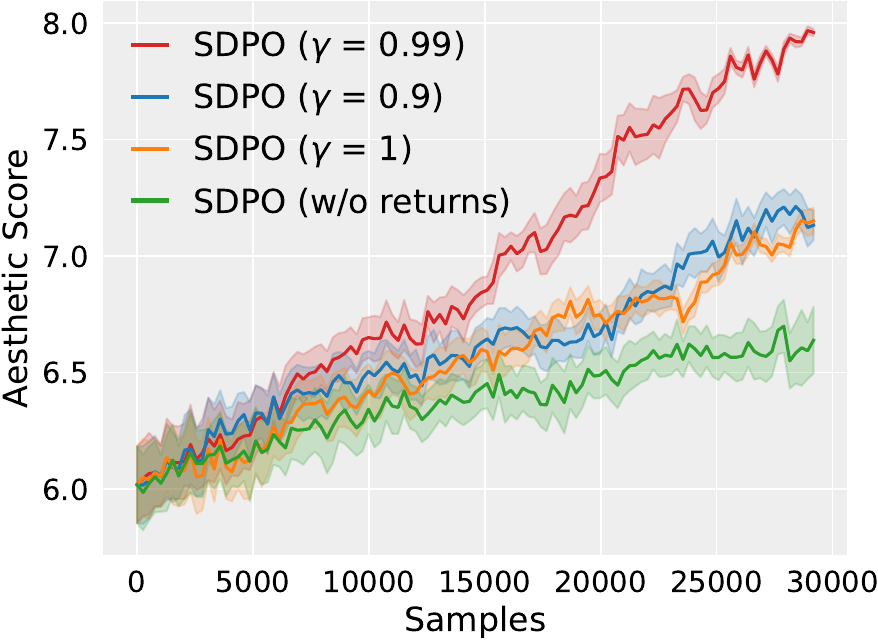}
    \caption{\textbf{Ablation study} on \textbf{dense reward prediction} (\textit{left \& middle}) and \textbf{discounted returns} (\textit{right}) for SDPO.}
    \label{fig:ablation1}
\end{figure*}

\subsection{Ablation Study}
\label{sec:4.3}

In this subsection, we conduct a series of ablation studies on SDPO's core components, providing comprehensive experimental analysis to validate their individual contributions to overall reward optimization performance, particularly in two-step denoising outputs. While reward functions like PickScore, HPSv2, and ImageReward benefit from explicit text-image alignment supervision, our ablation studies focus on the Aesthetic Score reward, which lacks such guidance and thus makes the reward optimization task more challenging.

\textbf{Effect of dense reward prediction.} As detailed in Section~\ref{sec:3.3}, we use 3 reward queries per trajectory, latent similarity, and an adaptively selected anchor to efficiently predict dense rewards. To examine the effectiveness of this approach, we compare SDPO with variants that employ alternative strategies for dense reward prediction, including:
\begin{itemize}
\item \textit{3 queries \& similarity \& adaptive anchor:} Original setup.
\item \textit{3 queries \& similarity \& random anchor:} In this variant, we retain the 3 reward queries per trajectory and the latent similarity computation, but select the anchor randomly instead of via adaptive criteria.
\item \textit{3 queries \& similarity \& fixed anchor:} In this setup, we again use 3 reward queries and latent similarity but set a fixed anchor at the 25th step for all trajectories. 
\item \textit{2 queries \& similarity:} Here, we reduce the number of reward queries from 3 to 2, retaining the latent similarity computation while eliminating the anchor selection.
\item \textit{2 queries \& interpolation:} In this method, we also use 2 reward queries but replace the latent similarity method with simple interpolation to approximate dense rewards.
\item \textit{1 query \& copy:} Here, we restrict reward queries to the final step, copying it to all intermediate steps and avoiding dense reward prediction—a common strategy used by most existing methods such as DDPO and REBEL.
\item \textit{Full-step queries:} This variant involves querying the reward function at every step of the trajectory, maximizing the reward density without any interpolation.
\end{itemize}

The left plot in Fig.~\ref{fig:ablation1} compares the sample efficiency of these variants by plotting the reward scores of two-step denoising outputs against training samples, while the middle plot further compares time efficiency by plotting reward scores against GPU hours. Notably, our original setup (\textit{3 queries \& similarity \& adaptive anchor}) consistently outperforms all variants in terms of both sample and time efficiency, underscoring the reliability and effectiveness of our dense reward prediction strategy. Interestingly, the \textit{full-step queries} variant, despite its maximal reward-query density, actually underperforms our method. This is likely due to the additional noise and variance introduced by direct per-step reward querying, which can destabilize policy updates. In contrast, our interpolation-based approach provides implicit regularization that smooths out such fluctuations while retaining essential reward dynamics, thereby facilitating more stable and efficient optimization. Moreover, the superior time efficiency of our method further highlights its practical advantages in delivering reliable dense rewards with minimal computational overhead.

\begin{figure*}[t]
    \centering
    \includegraphics[width=0.32\textwidth]{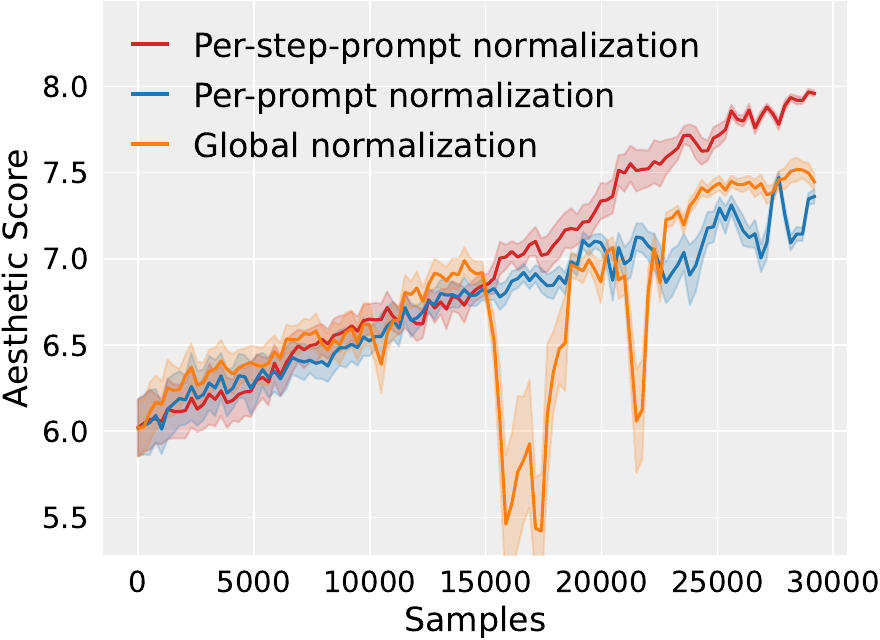}
    \includegraphics[width=0.32\textwidth]{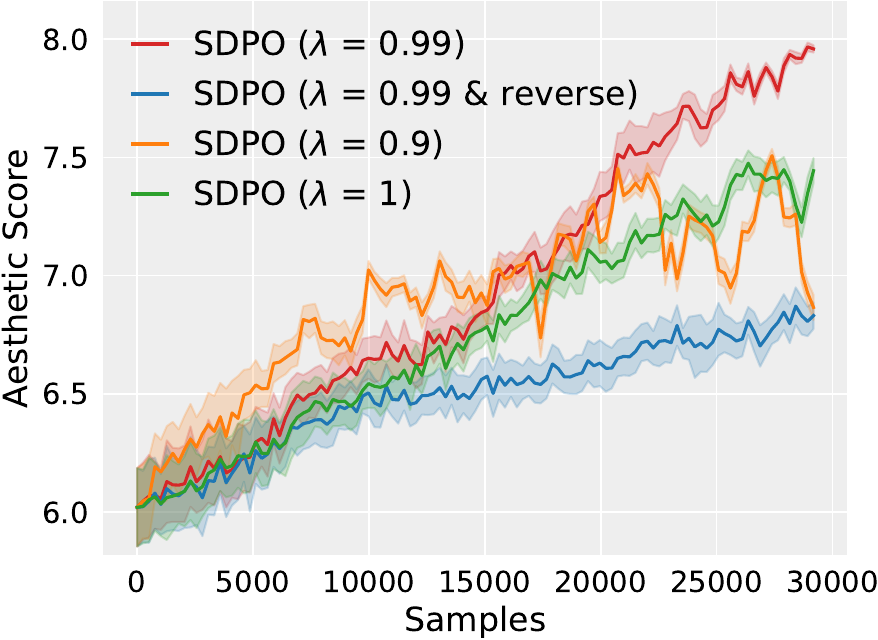}
    \includegraphics[width=0.33\textwidth]{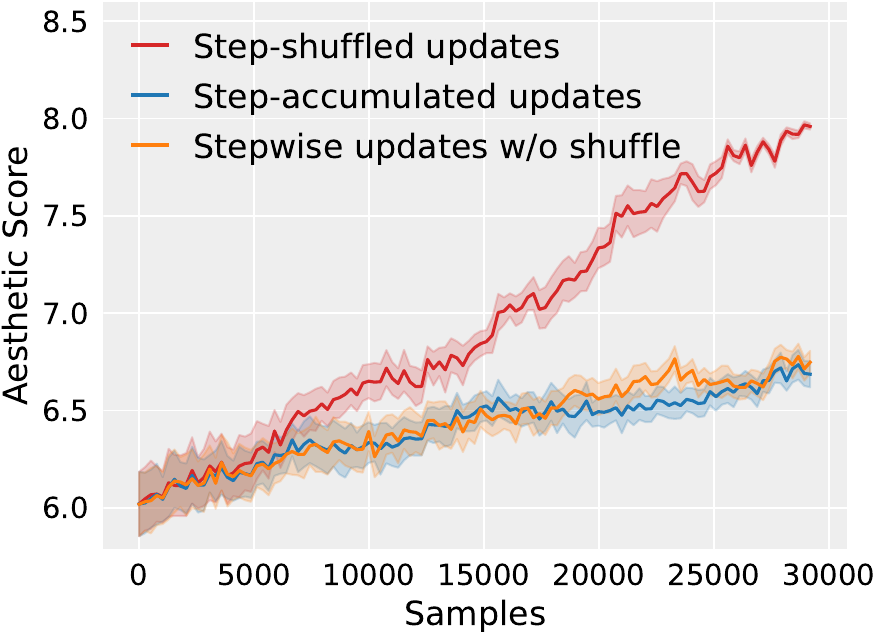}
    \caption{\textbf{Ablation study} on \textbf{return normalization} (\textit{left}), \textbf{importance weighting} (\textit{middle}), and \textbf{gradient updates} (\textit{right}) for SDPO.}
    \label{fig:ablation2}
\end{figure*}

\begin{table*}[t]
\caption{Comparison of similarity between predicted and directly queried rewards across different methods.}
\label{table:reward}
\centering
\begin{tabular}{l l c c c}
    \toprule
    \textbf{Reward Function} & \textbf{Method} & \textbf{Cosine Similarity $\uparrow$} & \textbf{L1 Distance $\downarrow$} & \textbf{L2 Distance $\downarrow$} \\
    \midrule
    \multirow{6}{*}{\textbf{Aesthetic Score}} & 1 query \& copy & $2\times10^{-8}$ & $11630$ & $131$ \\
    & 2 queries \& interpolation                                & $0.3368$ & $11228$ & $130$ \\
    & 2 queries \& similarity                                   & $0.3861$ & $9771$ & $125$ \\
    & 3 queries \& similarity \& fixed anchor                   & $0.3654$ & $9832$ & $127$ \\
    & 3 queries \& similarity \& random anchor                  & $0.3561$ & $9744$ & $128$ \\
    \rowcolor{blue!10}
    & 3 queries \& similarity \& adaptive anchor (Ours)         & $\mathbf{0.4463}$ & $\mathbf{9073}$ & $\mathbf{119}$ \\
    \midrule
    \multirow{6}{*}{\textbf{ImageReward}} & 1 query \& copy     & $2\times10^{-8}$ & $11262$ & $130$ \\
    & 2 queries \& interpolation                                & $0.3901$ & $10561$ & $125$ \\
    & 2 queries \& similarity                                   & $0.4078$ & $9340$ & $123$ \\
    & 3 queries \& similarity \& fixed anchor                   & $0.3611$ & $9423$ & $128$ \\
    & 3 queries \& similarity \& random anchor                  & $0.3924$ & $9337$ & $125$ \\
    \rowcolor{blue!10}
    & 3 queries \& similarity \& adaptive anchor (Ours)         & $\mathbf{0.4683}$ & $\mathbf{8598}$ & $\mathbf{117}$ \\
    \midrule
    \multirow{6}{*}{\textbf{HPSv2}} & 1 query \& copy           & $3\times10^{-8}$ & $11411$ & $129$ \\
    & 2 queries \& interpolation                                & $0.3696$ & $10798$ & $127$ \\
    & 2 queries \& similarity                                   & $0.3676$ & $9915$ & $127$ \\
    & 3 queries \& similarity \& fixed anchor                   & $0.3253$ & $9886$ & $131$ \\
    & 3 queries \& similarity \& random anchor                  & $0.3648$ & $9955$ & $128$ \\
    \rowcolor{blue!10}
    & 3 queries \& similarity \& adaptive anchor (Ours)         & $\mathbf{0.4549}$ & $\mathbf{9009}$ & $\mathbf{118}$ \\
    \midrule
    \multirow{6}{*}{\textbf{PickScore}} & 1 query \& copy       & $3\times10^{-8}$ & $11540$ & $131$ \\
    & 2 queries \& interpolation                                & $0.4307$ & $10237$ & $121$ \\
    & 2 queries \& similarity                                   & $0.4117$ & $9318$ & $123$ \\
    & 3 queries \& similarity \& fixed anchor                   & $0.3933$ & $9368$ & $125$ \\
    & 3 queries \& similarity \& random anchor                  & $0.3846$ & $9324$ & $126$ \\
    \rowcolor{blue!10}
    & 3 queries \& similarity \& adaptive anchor (Ours)         & $\mathbf{0.4853}$ & $\mathbf{8657}$ & $\mathbf{115}$ \\
    \bottomrule
\end{tabular}
\end{table*}

\textbf{Effect of discounted returns.} We then evaluate SDPO with different discount factors: $\gamma = 0.99$, $0.9$, and $1$, along with a variant that directly uses rewards without returns. As shown in the right plot of Fig.~\ref{fig:ablation1}, SDPO with $\gamma = 0.99$ achieves the best performance in sample efficiency.

\textbf{Effect of per-step-prompt normalization.} We also compare the per-step-prompt normalization strategy in SDPO with two alternatives: (1) \textit{per-prompt normalization}, which normalizes rewards on a per-prompt basis, and (2) \textit{global normalization}, which simply normalizes rewards across all samples. The left plot in Fig.~\ref{fig:ablation2} shows both alternatives underperform our per-step-prompt normalization. However, the per-prompt normalization variant still outperforms baseline methods, which also use per-prompt normalization, indicating that our dense reward strategy is reliable on its own.

\textbf{Effect of temporal importance weighting.} We further analyze the impact of temporal importance weighting in SDPO with different decay factors: $\lambda = 0.99$, $0.9$, and $1$, as well as a variant that applies weights in a \textit{reverse} order. The middle plot in Fig.~\ref{fig:ablation2} shows that SDPO with $\lambda = 0.99$ yields the best performance, while removing weighting ($\lambda = 1$) or applying reverse weighting degrades performance. Nonetheless, the variant with no weighting still outperforms baseline methods, further confirming the reliability of our dense reward strategy.

\textbf{Effect of step-shuffled gradient updates.} Finally, we evaluate the impact of replacing step-shuffled updates in SDPO with two alternatives: (1) \textit{step-accumulated updates} and (2) \textit{stepwise updates without shuffling}. As shown in the right plot in Fig.~\ref{fig:ablation2}, step-accumulated updates, which accumulate gradients across all denoising steps before a single model update, reduce sample efficiency compared to our step-shuffled updates. Additionally, stepwise updates performed sequentially without shuffling the step order also exhibit reduced sample efficiency, due to overfitting to the fixed step order.

\subsection{Analysis of Dense Reward Prediction Accuracy}
\label{sec:4.4}

To further substantiate the effectiveness of our dense reward prediction strategy, we compare dense rewards predicted by different strategies (from the ablation study in Section~IV-C) against ground-truth rewards directly queried from target reward functions along full trajectories. Specifically, we evaluate the similarity between predicted and queried rewards using cosine similarity as well as L1 and L2 distances. To address scale differences among reward functions, all rewards are normalized to the range [0, 1] before computing these metrics. As summarized in Table~\ref{table:reward}, our method consistently achieves the highest cosine similarity and lowest L1/L2 distances across all evaluated reward functions compared to alternative strategies. This demonstrates the effectiveness of our dense reward prediction strategy in capturing representative and reliable reward dynamics with limited reward queries.

\begin{figure*}[t]
    \centering
    \includegraphics[width=0.32\textwidth]{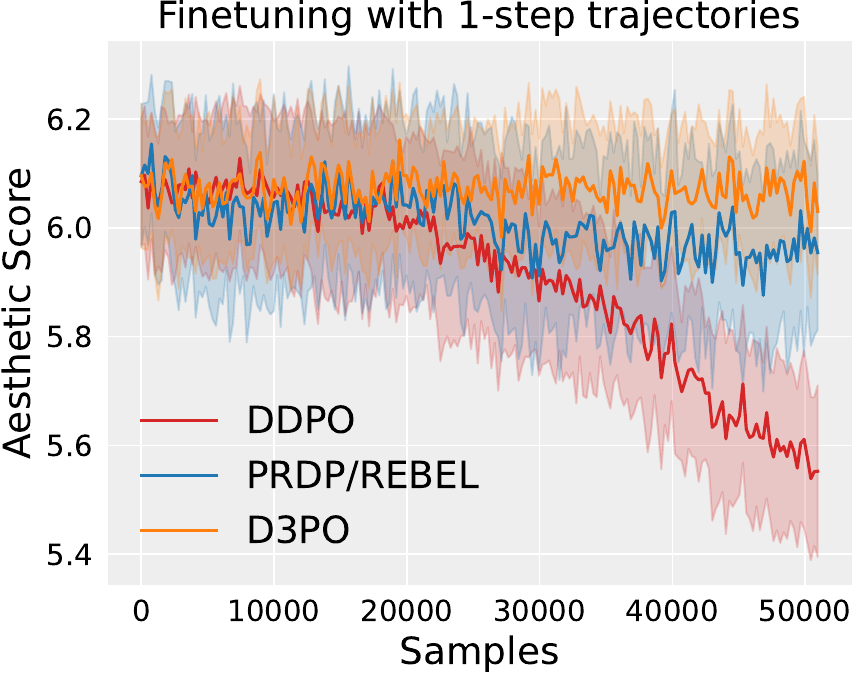}
    \hfill
    \includegraphics[width=0.32\textwidth]{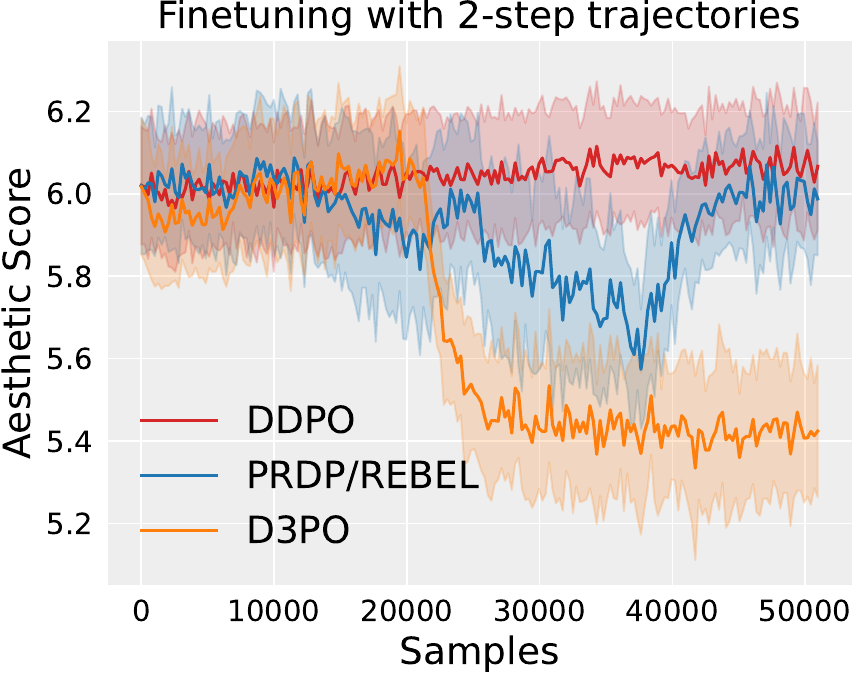}
    \hfill
    \includegraphics[width=0.32\textwidth]{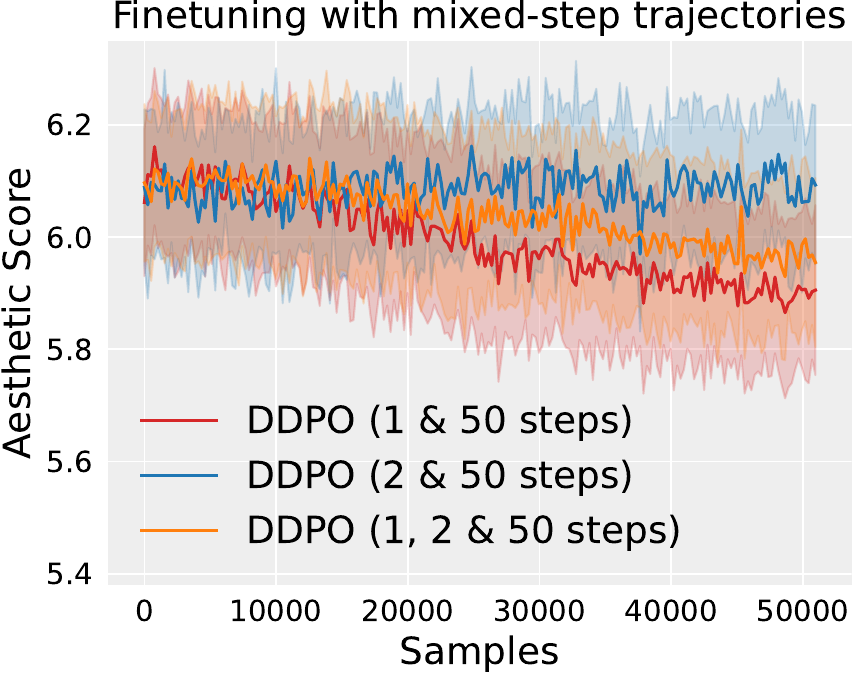}
    \caption{\textbf{Analysis of the instability of existing methods in finetuning with few-step or mixed-step trajectories.} We present the reward optimization curves of DDPO, PRDP/REBEL, and D3PO during finetuning SD-Turbo with \textbf{1-step} (\textit{left}) and \textbf{2-step} (\textit{middle}) denoising trajectories, as well as the reward optimization of DDPO during finetuning with \textbf{mixed-step} trajectories (\textit{right}).}
    \label{fig:instab}
\end{figure*}

\begin{figure*}[t]
    \centering
    \includegraphics[width=0.33\textwidth]{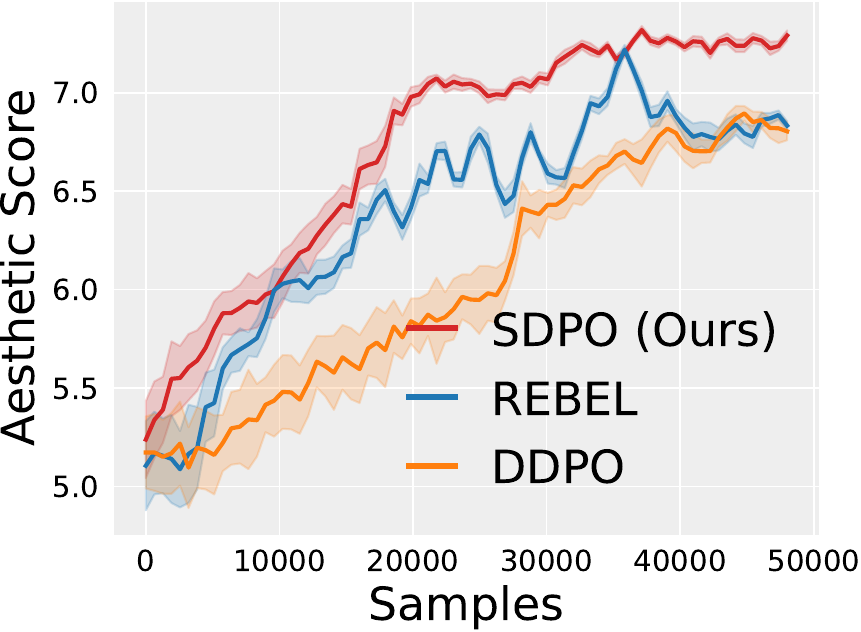}
    \hfill
    \includegraphics[width=0.32\textwidth]{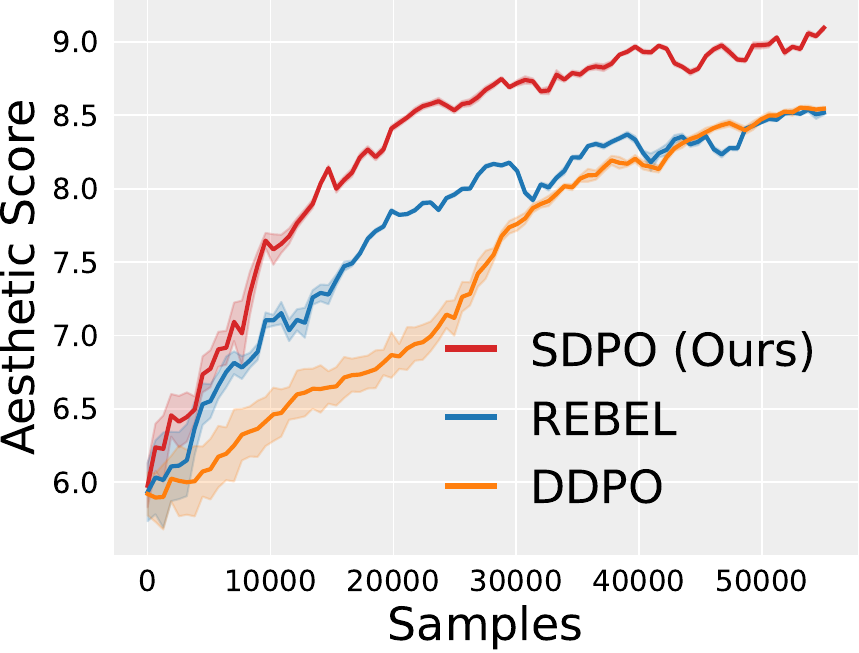}
    \hfill
    \includegraphics[width=0.32\textwidth]{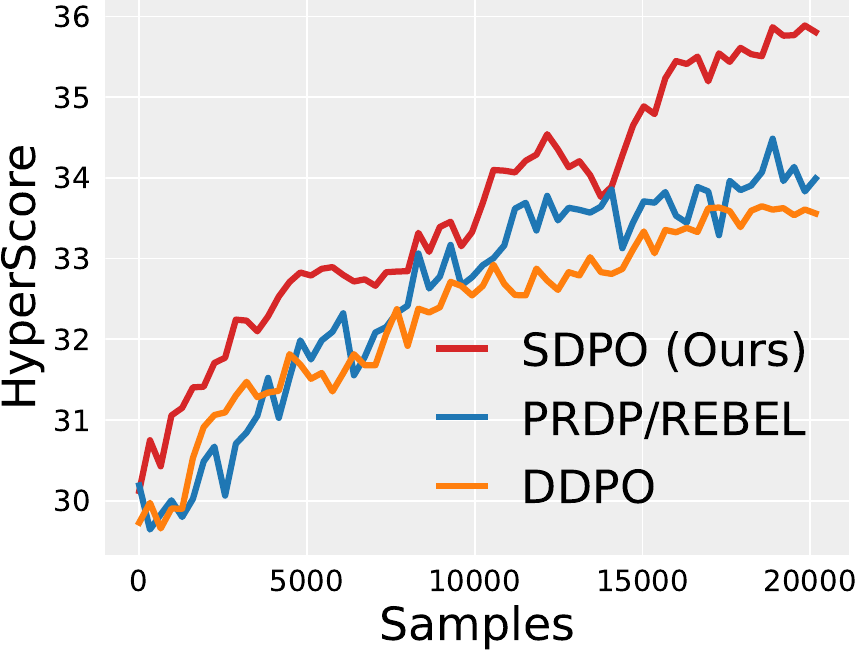}
    \caption{\textbf{Reward curves} for \textbf{latent consistency model} (\textit{left}: 1-step sampling; \textit{middle}: 8-step sampling) and \textbf{text-to-multiview diffusion model} (\textit{right}).}
    \label{fig:extension}
\end{figure*}

\subsection{Instability of Existing Methods in Finetuning with Few-Step and Mixed-Step Trajectories}
\label{sec:4.5}

As discussed in Section~\ref{sec:3.1}, we evaluate the training instability of existing methods (DDPO, PRDP/REBEL, and D3PO) when finetuning SD-Turbo with extremely few-step trajectories (e.g., 1 or 2 steps). The left and middle plots of Fig.~\ref{fig:instab} illustrate the reward optimization curves of these methods during finetuning with 1-step and 2-step denoising trajectories, respectively. The results reveal significant instability in reward optimization across all methods, characterized by large fluctuations and occasional collapses in reward scores over training iterations. This instability arises from the diminished quality and diversity of low-step samples, which lack sufficient information for the policy to distinguish subtle behaviors or assess long-term outcomes accurately.

Moreover, the right plot of Fig.~\ref{fig:instab} further displays the reward optimization curves of DDPO when finetuning with mixed-step trajectories. Here, we observe that DDPO still suffers from similar instability, as the mixed-step trajectories introduce additional variance and complexity into the training process. These findings underscore the challenges faced by existing methods in effectively leveraging few-step and mixed-step trajectories for stable and efficient reward optimization. In contrast to these methods, which rely on individual trajectories of varying lengths, our SDPO employs dual-state sampling within uniform-length trajectories to generate intermediate clean states that emulate outputs from varying-length trajectories. By incorporating this design with dense reward feedback, SDPO leverages consistent and informative training signals, thereby facilitating a balanced, stable optimization process.

\subsection{Extension to Latent Consistency Models}
\label{sec:4.6}

We further conduct extensive experiments on finetuning consistency models with our SDPO. To achieve this, we utilize the RLCM framework~\cite{rlcm}, which applies both DDPO and REBEL for finetuning a latent consistency model (LCM)~\cite{lcm} distilled from Dreamshaper v7~\cite{dreamshaper7}. As shown in Fig.~\ref{fig:extension}, SDPO achieves superior sample efficiency over DDPO and REBEL in finetuning the LCM on Aesthetic Score, even when evaluated under the most challenging 1-step sampling setting, further validating its effectiveness in extremely low-step regimes.

\subsection{Extension to Text-to-Multiview Diffusion Models}
\label{sec:4.7}

We also evaluate SDPO on the multiview image generation task. Specifically, we employ LCM-SDXL~\cite{lcmsdxl} with the MV-Adapter~\cite{mvadapter} as the base model and the HyperScore metric from the MATE-3D benchmark~\cite{mate3d} as the reward function. As illustrated in Fig.~\ref{fig:extension} (right), SDPO consistently outperforms competing methods in finetuning the text-to-multiview diffusion model on HyperScore.

\section{Related Work}
\label{sec:5}

\textbf{Diffusion model alignment.} Existing approaches for diffusion alignment~\cite{lee2023,hps,liu2024alignment} can be broadly classified into two categories: (1) reward-based methods, which use techniques such as policy gradient~\cite{dpok,ddpo,tdpo,zhang2024large,uehara2024feedback,lee2024parrot,cdpo,liu2025flow}, reward difference~\cite{prdp,rebel}, or reward backpropagation through sampling~\cite{draft,alignprop,prabhudesai2024video,wu2024deep} to optimize the model based on explicit reward functions; and (2) preference-based methods~\cite{d3po,wallace2024diffusion,yang24a,li2024aligning,yuan2024self,spo,karthik2025scalable,croitoru2025curriculum}, which employ DPO-style objectives to directly align the model with data-driven preferences. Our work falls into the first category, focusing on aligning few-step diffusion models.

\textbf{Reward finetuning of few-step diffusion models.} Recent approaches~\cite{li2024reward,ren2024hyper} incorporate reward guidance into consistency distillation, while not directly applicable to other distilled few-step models. RLCM~\cite{rlcm} introduces an RL finetuning framework specifically for consistency models, which we use as a baseline in our extensive analysis of consistency models. In contrast, we extend RL finetuning to a wider range of few-step diffusion models and address the challenge of optimizing extremely low-step samples.

\textbf{Dense feedback for diffusion models.}
Recent studies have explored dense feedback in diffusion alignment. For RL finetuning, Zhang et al.~\cite{tdpo} highlight the importance of dense rewards to prevent overoptimization, and introduce a temporal critic to estimate dense rewards. For preference learning, Yang et al.~\cite{yang24a} apply temporal discounting in a DPO-style objective to distinguish preferences at each step, while Liang et al.~\cite{spo} build a step-aware preference model to provide stepwise preference. In contrast, our work incorporates dense rewards into RL finetuning of few-step diffusion models without requiring additional model training.

\section{Conclusion}
\label{sec:6}

This paper presents SDPO, a novel framework for RL finetuning of few-step diffusion models. By unifying dual-state trajectory sampling with dense reward feedback in a stepwise advantage difference learning paradigm, SDPO enables stable and efficient reward optimization, especially in extremely low-step regimes where existing methods falter. Future work may explore extending SDPO to a wider range of reward functions and state-of-the-art models, as well as integrating it with advanced RL finetuning techniques such as Flow-GRPO~\cite{liu2025flow}, to further evaluate and enhance its performance and scalability across diverse diffusion model alignment scenarios.

\section*{Acknowledgments}
This work was supported in part by the National Natural Science Foundation of China under Grant 62225113, Grant U23A20318, and Grant 62276195, in part by the Science and Technology Major Project of Hubei Province under Grant 2025BCB026 and Grant 2024BAB046, in part by the Foundation for Innovative Research Groups of Hubei Province under Grant 2024AFA017.
The work of Li Shen was supported in part by NSFC under Grant 62576364, in part by Shenzhen Basic Research Project (Natural Science Foundation) Basic Research Key Project under Grant JCYJ20241202124430041, in part by CCF-Tencent Rhino-Bird Open Research Fund under Grant CCF-Tencent RAGR20250114 and Grant Tencent JR2025TEG002.
Dr Tao's research is partially supported by NTU RSR and Start Up Grants.
This work was also supported by WHU-Kingsoft Joint Lab.
The numerical calculations in this paper have been done on the supercomputing system in the Supercomputing Center of Wuhan University.

\bibliographystyle{IEEEtran}
\bibliography{TPAMI-2025-07-2236.R2_Luo}

\begin{IEEEbiography}[{\includegraphics[height=1.25in,clip,keepaspectratio]{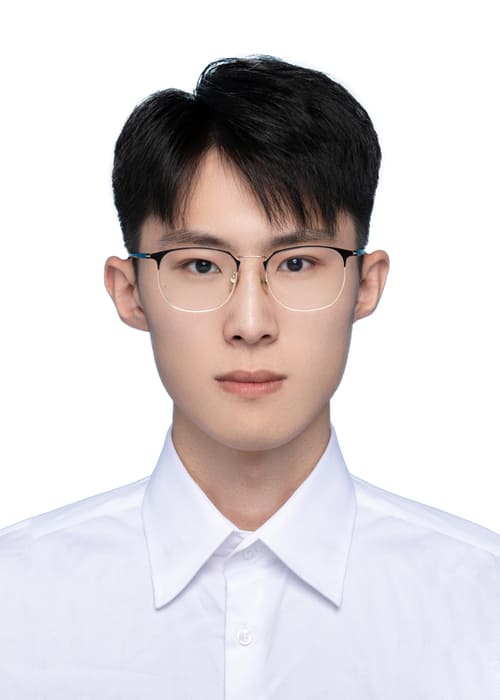}}]{Ziyi Zhang} is currently working toward the PhD degree with the School of Computer Science, Wuhan University, China, supervised by Prof. Dr. Yong Luo. His research interests currently focus on generative models (particularly diffusion models) and reinforcement learning from human feedback. He has published several papers in peer-reviewed top-tier journals and conferences, including IEEE TPAMI, ICML, IJCAI, and IEEE TMM. His personal homepage is at \href{https://ziyizhang27.github.io}{https://ziyizhang27.github.io}.
\end{IEEEbiography}

\begin{IEEEbiography}[{\includegraphics[height=1.25in,clip,keepaspectratio]{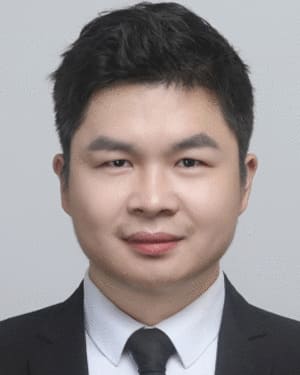}}]{Li Shen} is currently an associate professor at Sun Yat-sen University. Previously, he was a research scientist at JD Explore Academy, Beijing, and a senior researcher at Tencent AI Lab, Shenzhen. He received his bachelor's degree and Ph.D. from the School of Mathematics, South China University of Technology. His research interests include efficient deep learning, efficient reinforcement learning, optimization and deep learning theory. He has served as the senior program committee for AAAI and area chairs for ICML, NeurIPS, ICLR, CVPR, and ACM MM. He is also the associate editor for IEEE Transactions on Pattern Analysis and Machine Intelligence, IEEE Transactions on Knowledge and Data Engineering, and IEEE Transactions on Multimedia.
\end{IEEEbiography}

\begin{IEEEbiography}[{\includegraphics[height=1.25in,clip,keepaspectratio]{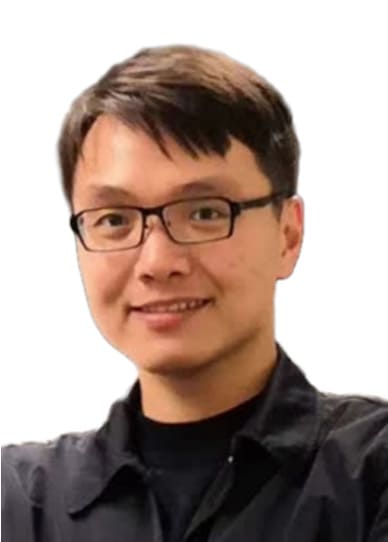}}]{Sen Zhang} received the Ph.D. degree from the School of Computer Science at the University of Sydney. He is currently a machine learning engineer at TikTok, ByteDance, Sydney. Previously, he was a postdoctoral research assistant at the University of Sydney. His research interests include computer vision, vSLAM, and foundation models. He has published several papers in top-tier conferences and journals, including IEEE TPAMI, ECCV, IJCV, ICRA, ICML, ICLR, and ACM Multimedia.
\end{IEEEbiography}

\begin{IEEEbiography}[{\includegraphics[height=1.25in,clip,keepaspectratio]{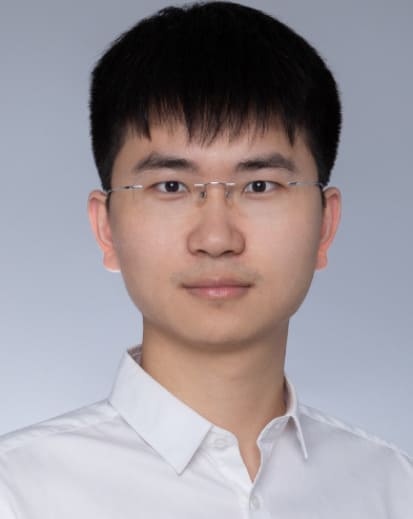}}]{Deheng Ye} received the Ph.D. degree from the School of Computer Science and Engineering, Nanyang Technological University, Singapore, in 2016. He is currently the Director of AI Applications at Tencent, Shenzhen, China, where he spearheads a dynamic team of engineers and researchers in the development of large-scale machine learning platforms and intelligent AI agents. His leadership has been instrumental in innovating and integrating pioneering AI technologies into real-world applications, with online AI service requests reaching billions per day worldwide. Apart from industry-level projects, he has authored/co-authored more than 50 academic papers to date. He also regularly serves as PC/SPC for leading venues such as NeurIPS, ICML, ICLR, AAAI, IJCAI, and so on.
\end{IEEEbiography}

\begin{IEEEbiography}[{\includegraphics[height=1.25in,clip,keepaspectratio]{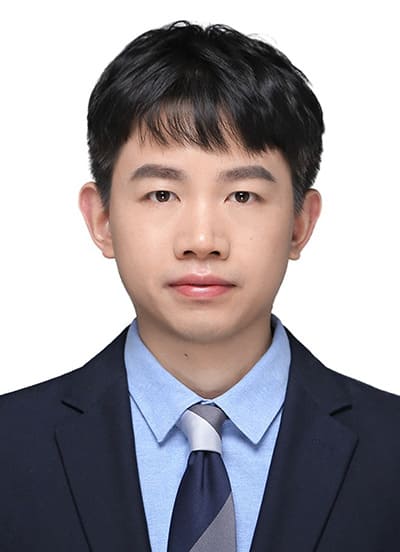}}]{Yong Luo} is currently a Professor with the School of Computer Science, Wuhan University, China. His research interests are primarily on machine learning and data mining with applications to visual information understanding and analysis. He has authored or co-authored over 100 papers in top journals and prestigious conferences including Nature Machine Intelligence, Nature Communications, IEEE T-PAMI and IJCV. He is serving on editorial board for IEEE T-MM and The Innovation Informatics. He received the IEEE Globecom 2016 Best Paper Award, and was nominated as the IJCAI 2017 Distinguished Best Paper Award. He is also a co-recipient of the IEEE TMM 2023, IEEE ICME 2019 and IEEE VCIP 2019 Best Paper Awards.
\end{IEEEbiography}

\begin{IEEEbiography}[{\includegraphics[height=1.25in,clip,keepaspectratio]{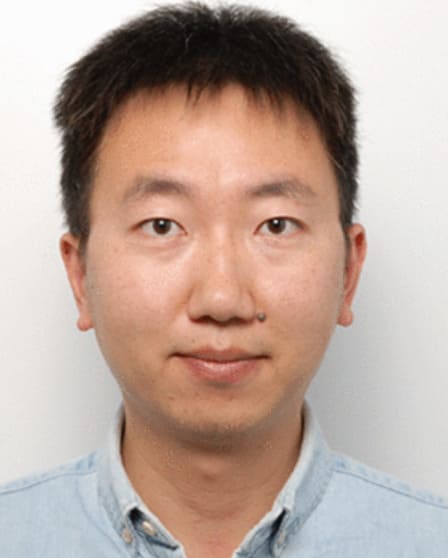}}]{Miaojing Shi} received the Ph.D. degree from Peking University in 2015. He also engaged with a joint PhD program with the University of Oxford and INRIA Rennes for a year. He held a postdoctoral position with the University of Edinburgh and was a research scientist with INRIA Rennes. Between 2020 and 2022, he was a lecturer with the Department of Informatics, King's College London. Since 2023, he has become a full professor with Tongji University and a visiting senior lecturer with King's. He has authored or coauthored more than 50 publications. His current research focus is on visual learning and understanding, with multi-task and limited supervision.
\end{IEEEbiography}

\begin{IEEEbiography}[{\includegraphics[height=1.25in,clip,keepaspectratio]{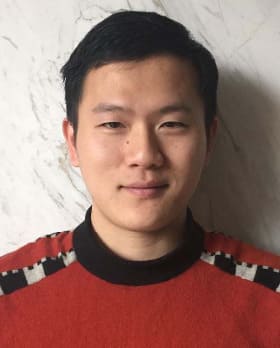}}]{Dongjing Shan} is currently an Associate Professor with the School of Medical Information and Engineering, Southwest Medical University, Luzhou, Sichuan, China. Previously, he was a Researcher with the Laboratory of Intelligent Information Processing, Army Engineering University, Chongqing, China. He received the B.Eng. degree in computer science from Beijing Institute of Technology, Beijing, China, in 2008, the M.S. degree in information science and technology from Peking University, Beijing, in 2013, and the Ph.D. degree in signal processing from the Speech Processing Laboratory, Army Engineering University, Nanjing, China, in 2020. His research interests include machine learning and deep learning with applications to image processing and speech recognition.
\end{IEEEbiography}

\begin{IEEEbiography}[{\includegraphics[height=1.25in,clip,keepaspectratio]{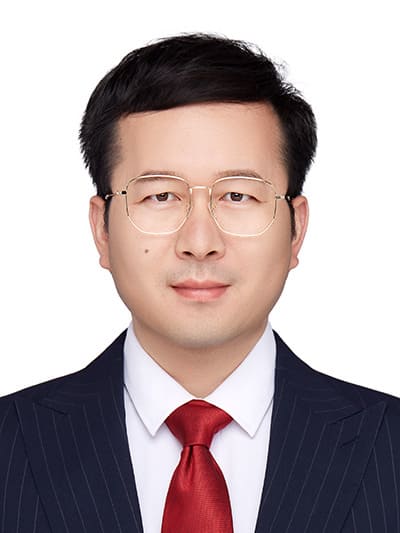}}]{Bo Du} received the Ph.D. degree in photogrammetry and remote sensing from the State Key Laboratory of Information Engineering in Surveying, Mapping and Remote Sensing, Wuhan University, Wuhan, China, in 2010. He is currently a Professor with the School of Computer Science and the Institute of Artificial Intelligence, Wuhan University. He is also the Director of the National Engineering Research Center for Multimedia and Software, Wuhan University. He has published more than 100 research articles in IEEE Transactions on Pattern Analysis and Machine Intelligence (TPAMI), IEEE Transactions on Image Processing (TIP), IEEE Transactions on Cybernetics (TCYB), IEEE Transactions on Geoscience and Remote Sensing (TGRS), IEEE Journal of Selected Topics in Applied Earth Observations and Remote Sensing (JSTARS), and IEEE Geoscience and Remote Sensing Letters (GRSL). His research interests include pattern recognition, hyperspectral image processing, and signal processing. 
He received the IEEE Geoscience and Remote Sensing Society (GRSS) 2020 Transactions Prize Paper Award, the IJCAI Distinguished Paper Prize, the IEEE Data Fusion Contest Champion, and the IEEE Workshop on Hyperspectral Image and Signal Processing Best Paper Award in 2018. He has served as the Area Chair for the International Conference on Pattern Recognition (ICPR). He serves as an Associate Editor for Neural Networks, Pattern Recognition, and Neurocomputing.
\end{IEEEbiography}

\begin{IEEEbiography}[{\includegraphics[height=1.25in,clip,keepaspectratio]{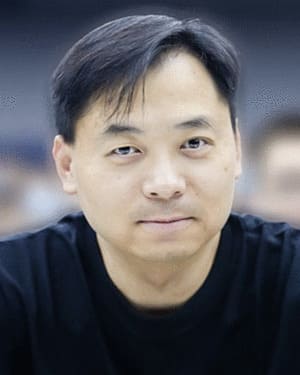}}]{Dacheng Tao} is currently a Distinguished University Professor and the Inaugural Director of the Generative AI Lab in the College of Computing and Data Science at Nanyang Technological University. He was an Australian Laureate Fellow and the founding director of the Sydney AI Centre in the University of Sydney, the inaugural director of JD Explore Academy and senior vice president in JD.com, and the chief AI scientist in UBTECH Robotics. He mainly applies statistics and mathematics to artificial intelligence, and his research is detailed in one monograph and over 300 publications. His publications have been cited over 140K times and he has an h-index 180+ in Google Scholar. He received the 2015 and 2020 Australian Eureka Prize, the 2018 IEEE ICDM Research Contributions Award, 2020 research super star by The Australian, the 2019 Diploma of The Polish Neural Network Society, and the 2021 IEEE Computer Society McCluskey Technical Achievement Award. He is a Fellow of the Australian Academy of Science, ACM and IEEE.
\end{IEEEbiography}

\vfill

\includepdf[pages=-]{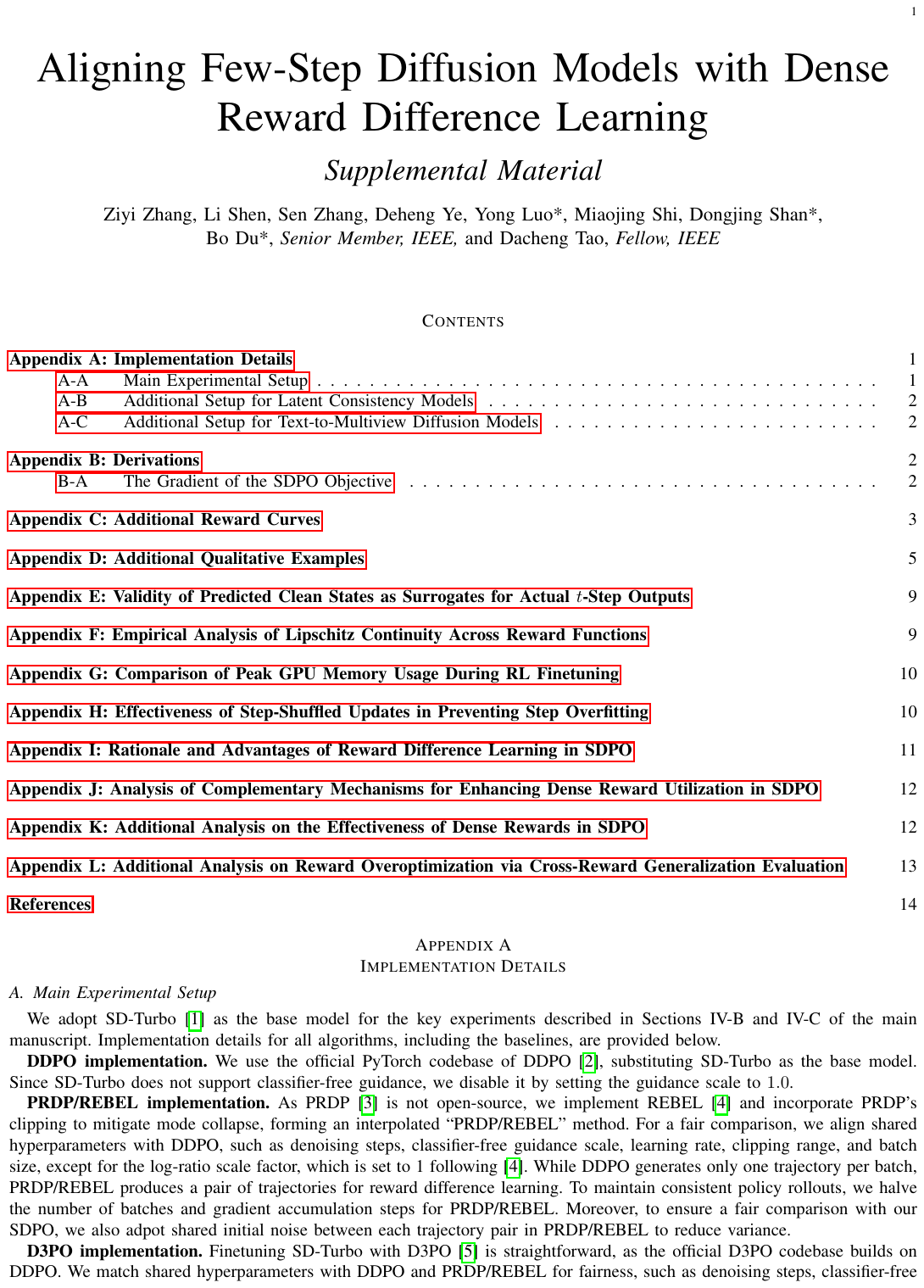}

\end{document}